%% file: 0_main.tex
\documentclass[11pt,a4paper]{article}
\usepackage[hyperref]{emnlp2020}
\usepackage{times}
\usepackage{latexsym}

\usepackage{microtype}

\usepackage{amsmath}
\usepackage{amssymb}
\usepackage{booktabs}
\usepackage{graphicx}
\usepackage{hyperref}
\usepackage{paralist}
\usepackage{times}
\usepackage{url}
\usepackage{xspace}
\usepackage{diagbox}
\usepackage{multirow, makecell}
\usepackage{subcaption}
\usepackage{tikz}
\usepackage{xcolor}
\usepackage{soul}
\usepackage{amssymb}
\usepackage{pifont}
\usepackage{array} 

\aclfinalcopy 
\def\aclpaperid{2137} 

\newcommand\gpt{{\small\textsc{GPT-2}}\xspace}
\newcommand\vqa{\textsc{vqa}\xspace}
\newcommand\vqae{\textsc{vqa-e}\xspace}
\newcommand\esnlive{\textsc{e-snli-ve}\xspace}
\newcommand\esnli{\textsc{e-snli}\xspace}
\newcommand\vcr{\textsc{vcr}\xspace}

\newcommand\modelname{\textsc{\small Rationale}$^{\textsc{\small vt}}$ \textsc{\small Transformer}\xspace}
\newcommand\uniform{\textsc{uniform}\xspace}
\newcommand\hybrid{\textsc{hybrid}\xspace}
\newcommand\viscomet{{\small\textsc{VisComet}}\xspace}
\newcommand\visualcomet{{\small\textsc{VisualComet}}\xspace}

\definecolor{lightblue}{RGB}{186, 202, 255}
\definecolor{lightred}{RGB}{255, 188, 186}
\definecolor{lightgreen}{RGB}{209, 255, 186}

\newcommand\sect[1]{\S\ref{#1}}

\title{
Natural Language Rationales with Full-Stack Visual Reasoning: \\ 
From Pixels to Semantic Frames to Commonsense Graphs
}

\author{
	Ana Marasovi\'{c}$^{\dagger\diamondsuit}$ \quad
	Chandra Bhagavatula$^{\dagger}$ \quad 
    Jae Sung Park$^{\diamondsuit\dagger}$  \quad \\
	\bf Ronan Le Bras$^\dagger$ \quad
	Noah A. Smith$^{\diamondsuit\dagger}$ \quad
	Yejin Choi$^{\diamondsuit\dagger}$ \\\\
	$^\dagger$Allen Institute for Artificial Intelligence\\
	$^\diamondsuit$Paul G. Allen School of Computer Science \& Engineering, University of Washington \\
	Seattle, WA, USA\\
	{\tt \{anam,chandrab,jamesp,ronanlb\}@allenai.org} \\ 
	{\tt \{jspark96,nasmith,yejin\}@cs.washington.edu}
}

\date{}

\begin{document}
\maketitle

\begin{figure*}[!h]
\centering
\includegraphics[width=0.82\textwidth]{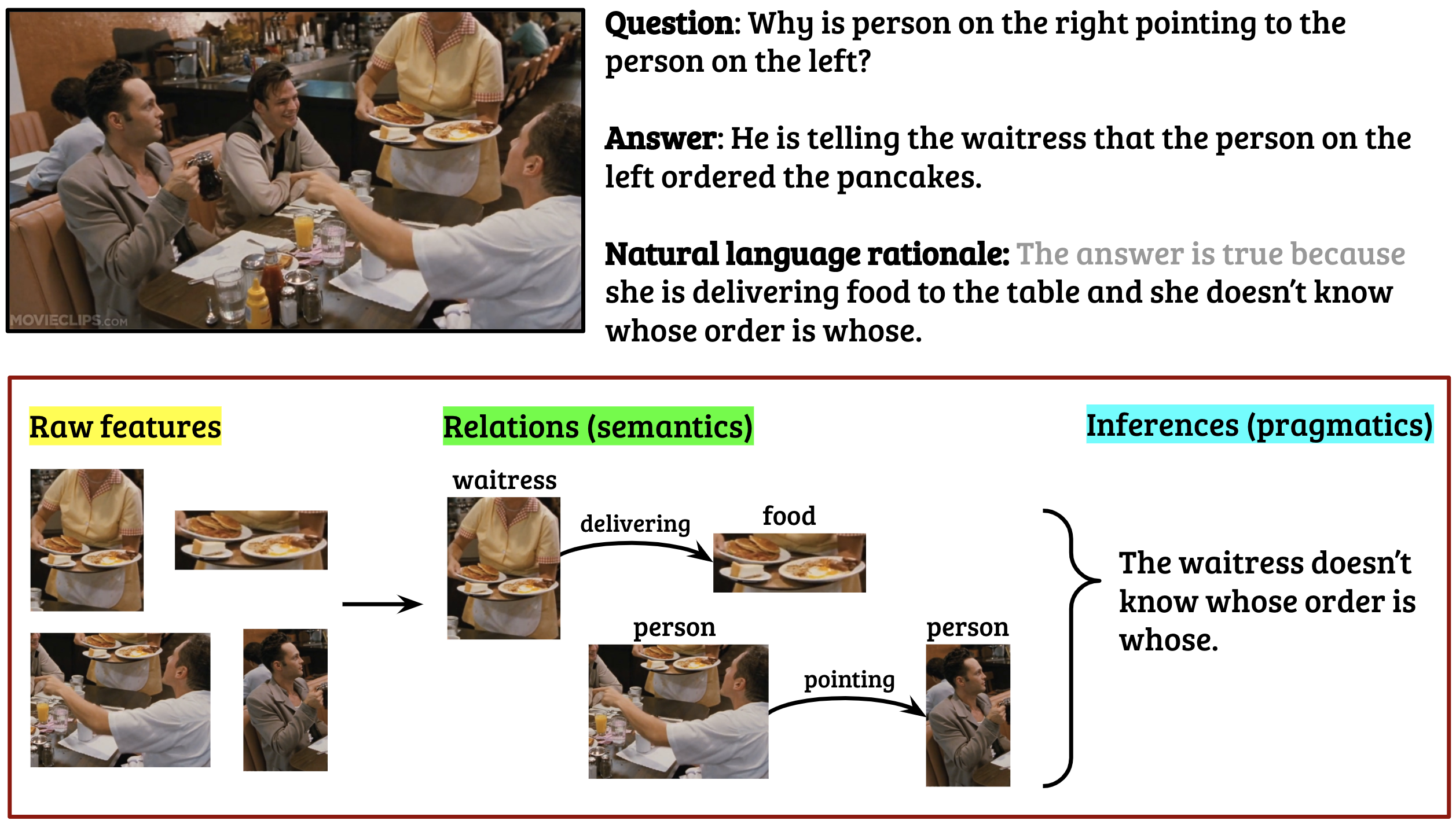}
      \caption{An illustrative example showing that explaining higher-level conceptual reasoning cannot be well conveyed only through the attribution of raw input features (individual pixels or words); we need natural language.} 
     \label{fig:motivation}
\end{figure*}

\begin{abstract} 
Natural language rationales could provide intuitive, higher-level explanations that are easily understandable by humans, complementing the more broadly studied lower-level explanations based on gradients or attention weights.  
We present the first study focused on generating natural language rationales across several complex visual reasoning tasks: visual commonsense reasoning, visual-textual entailment, and visual question answering. 
The key challenge of accurate rationalization is comprehensive image understanding at all levels: not just their explicit content at the pixel level, but their contextual contents at the semantic and pragmatic levels.  
We present \modelname, an integrated model that learns to generate free-text rationales by combining pretrained language models with  object recognition, grounded visual semantic frames, and visual commonsense graphs. 
Our experiments show that the base pretrained language model benefits from visual adaptation and that free-text rationalization is a promising research direction to complement model interpretability for complex visual-textual reasoning tasks. 

\end{abstract}

\input{1_intro}

\input{3_multimodal_generation}
\input{4_experiments}

\input{5_related_work}

\input{6_conclusions}

\section*{Acknowledgments}
The authors thank Sarah Pratt for her assistance with the grounded situation recognizer, Amandalynne Paullada, members of the AllenNLP team, and anonymous reviewers for helpful feedback. This research was supported in part by NSF (IIS1524371, IIS-1714566), DARPA under the CwC program through the ARO (W911NF-15-1-0543), DARPA under the MCS program through NIWC Pacific (N66001-19-2-4031), and gifts from Allen Institute for Artificial Intelligence.

\bibliography{emnlp2020}
\bibliographystyle{acl_natbib}
\clearpage
\appendix

\input{supplement}

\end{document}

%% file: 1_intro.tex
\section{Introduction}
\label{sec:intro}

Explanatory models based on natural language rationales could provide intuitive, higher-level explanations that are easily understandable by humans \cite{Miller2019ExplanationIA}. 
In Figure~\ref{fig:motivation}, for example, the natural language rationale given in free-text provides a much more informative and conceptually relevant explanation to the given QA problem compared to the non-linguistic explanations that are often provided as localized visual highlights on the image. 
The latter, while pertinent to what the vision component of the model was attending to, cannot provide the full scope of rationales for such complex reasoning tasks as illustrated in Figure~\ref{fig:motivation}. Indeed, explanations for higher-level conceptual reasoning can be best conveyed through natural language, as has been studied in literature on (visual) NLI \cite{Do2020eSNLIVE20CV, camburu_e_snli_2019}, (visual) QA \cite{wu-mooney-2019-faithful, rajani-etal-2019-explain}, arcade games \cite{Ehsan2019AutomatedRG}, fact checking \cite{atanasova-etal-2020-generating}, image classification \cite{Hendricks2018GroundingVE}, motivation prediction \cite{Vondrick2016PredictingMO}, algebraic word
problems \cite{ling-etal-2017-program}, and self-driving cars \cite{Kim2018TextualEF}. 

In this paper, we present the first focused study on generating natural language rationales across several complex visual reasoning tasks: visual commonsense reasoning, visual-textual entailment, and visual question answering. 
Our study aims to \textit{complement} the more broadly studied lower-level explanations such as attention weights and gradients in deep neural networks \cite[among others]{Simonyan2014DeepIC, Zhang2017TopDownNA, Montavon2018MethodsFI}.
Because free-text rationalization is a challenging research question, we assume the gold answer for a given instance is given and scope our investigation to justifying the gold answer.  

The key challenge in our study is that accurate rationalization requires comprehensive image understanding at all levels: not just their basic 
content at the pixel level (recognizing ``waitress'', ``pancakes'', ``people'' at the table in Figure \ref{fig:motivation}), but their contextual content at the semantic level (understanding the structural relations among objects and entities through action predicates such as ``delivering'' and ``pointing to'') as well as at the pragmatic level (understanding the ``intent'' of the pointing action is to tell the waitress who ordered the pancakes).  

We present \modelname, an integrated model that learns to generate free-text rationales by combining pretrained language models based on \gpt \cite{Radford2019LanguageMA} with visual features. Besides commonly used features derived from object detection (Fig.\ \ref{fig:object_detection}), we explore two new types of visual features to enrich base models with semantic and pragmatic knowledge: (i) visual semantic frames, i.e., the primary activity and entities engaged in it detected by a grounded situation recognizer \cite[Fig.\  \ref{fig:cv_tools_gsr};][]{Pratt2020GroundedSR}, and (ii) commonsense inferences inferred from an image and an optional event predicted from a visual commonsense graph \cite[Fig.\ \ref{fig:cv_tools_vcg};][]{Park2020VisualCG}.\footnote{Figures \ref{fig:object_detection}--\ref{fig:cv_tools_vcg} are taken and modified from \citet{Zellers2018FromRT}, \citet{Pratt2020GroundedSR}, and \citet{Park2020VisualCG}, respectively.} 

We report comprehensive experiments with careful analysis using three datasets with human rationales: (i) visual question answering in \textsc{vqa-e} \cite{Li2018VQAEEE}, (ii) visual-textual entailment in \textsc{e-snli-ve} \cite{Do2020eSNLIVE20CV}, and (iii) an answer justification subtask of visual commonsense reasoning in \vcr \cite{Zellers2018FromRT}. 
Our empirical findings demonstrate that while free-text rationalization remains a challenging task, newly emerging state-of-the-art models support rationale generation as a promising research direction to complement model interpretability for complex visual-textual reasoning tasks. 
In particular, we find that integration of richer semantic and pragmatic visual knowledge can lead to generating rationales with higher visual fidelity, especially for tasks that require higher-level concepts and richer background knowledge.

Our code, model weights, and the templates used for human evaluation  are publicly available.\footnote{\url{https://github.com/allenai/visual-reasoning-rationalization}}

\input{figs/vision_tools}

%% file: figs/vision_tools.tex
\begin{figure*}[t]
\centering
\begin{subfigure}[t]{.4\textwidth}
  \centering
  \includegraphics[width=\linewidth]{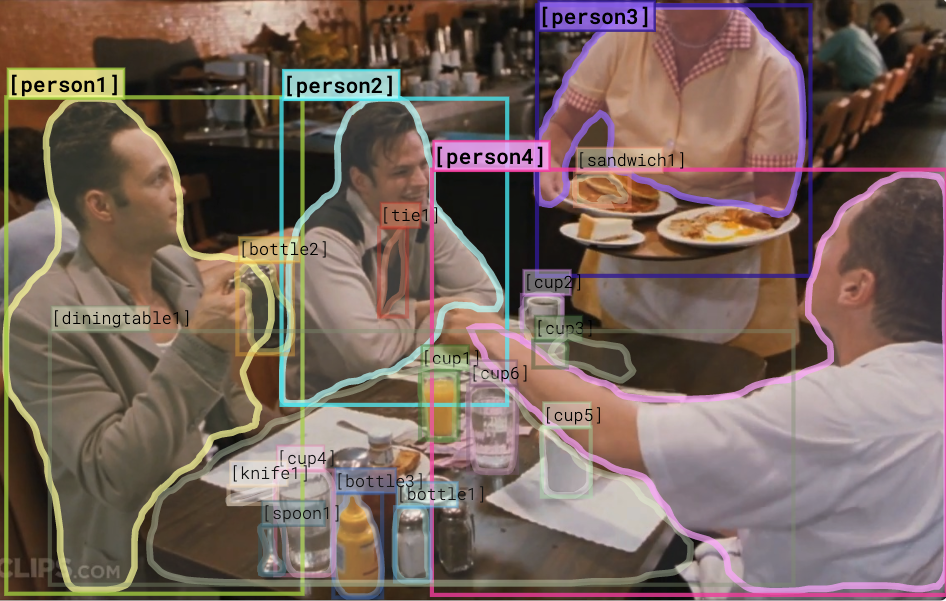}
  \caption{Object Detection}
  \label{fig:object_detection}
\end{subfigure}\text{ }
\begin{subfigure}[t]{.29\textwidth}
  \centering
  \includegraphics[width=0.92\linewidth]{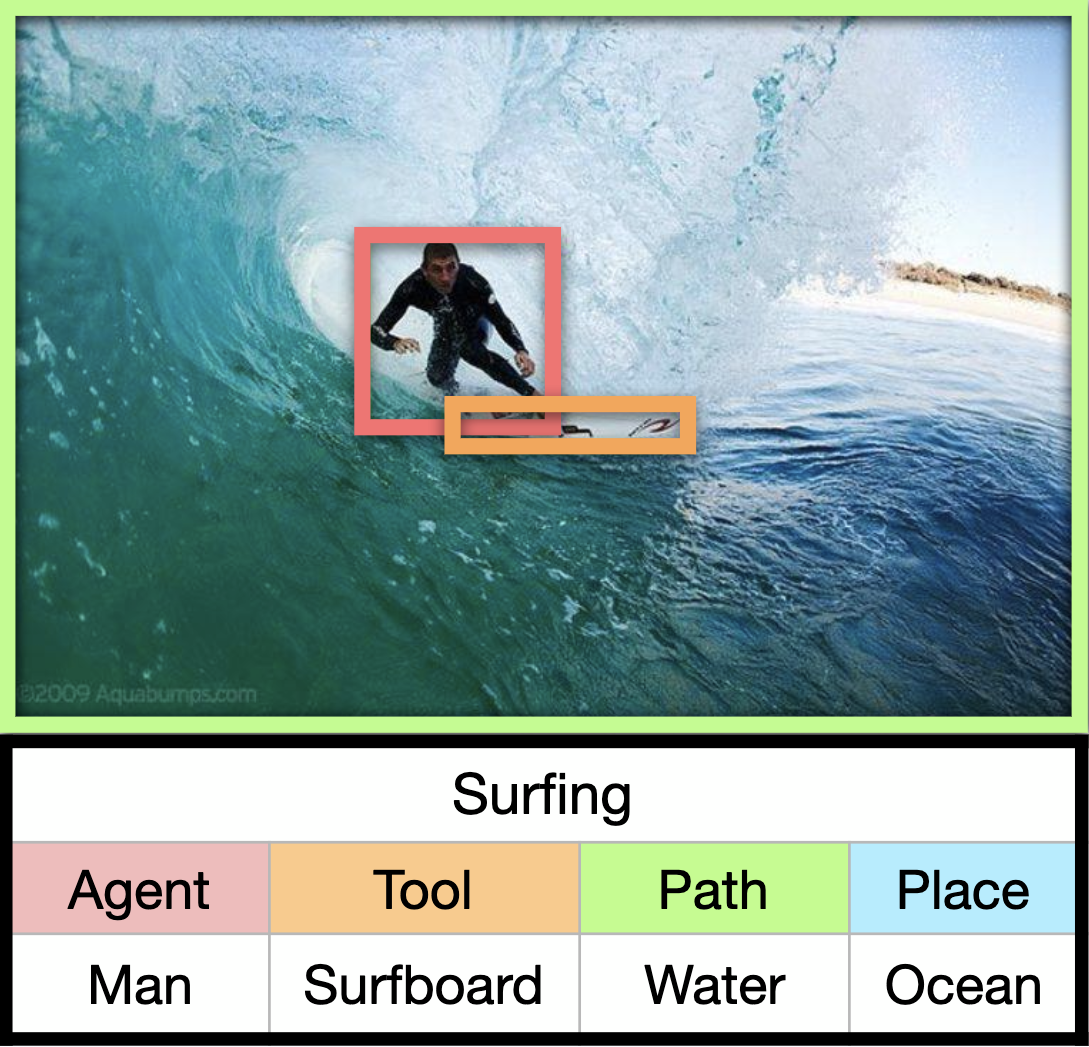}
  \caption{Grounded Situation Recognition}
  \label{fig:cv_tools_gsr}
\end{subfigure}\text{ }
\begin{subfigure}[t]{.27\textwidth}
  \centering
  \includegraphics[width=\linewidth]{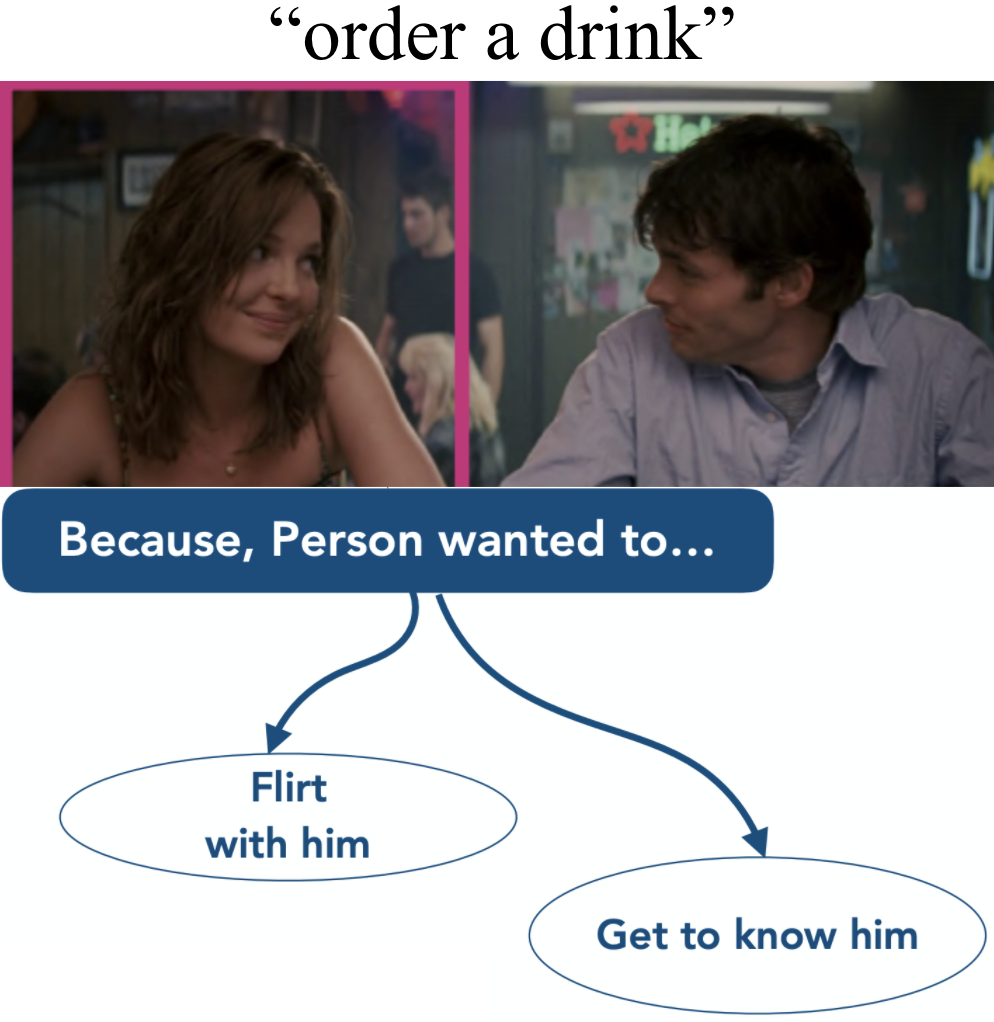}
  \caption{Visual Commonsense Graph}
  \label{fig:cv_tools_vcg}
\end{subfigure}
\caption{An illustration of outputs of external vision models that we use to visually adapt \gpt.}
\label{fig:cv_tools}
\end{figure*}

%% file: 3_multimodal_generation.tex
\input{tables/3_vision_models}
\input{tables/4_datasets}

\section{Rationale Generation with \textsc{Rationale}$^{\textsc{vt}}$ \textsc{Transformer}}
\label{sec:visioling_generation}

Our approach to visual-textual rationalization is based on augmenting \gpt's input with output of external vision models that enable different levels of visual understanding.

\subsection{Background: Conditional Text Generation}

The \gpt's backbone architecture can be described as the decoder-only Transformer \cite{vaswani2017attention} which is pretrained with the conventional language modeling (LM) likelihood objective.\footnote{Sometimes referred to as density estimation, or left-to-right or autoregressive LM \cite{Yang2019XLNetGA}.} This makes it more suitable for generation tasks compared to models trained with the masked LM objective \cite[BERT;][]{devlin-etal-2019-bert}.\footnote{See Appendix \sect{sec:appendix_gpt_details} for other details of \gpt.} 

We build on pretrained LMs because their capabilities make free-text rationalization of complex reasoning tasks conceivable. They strongly condition on the preceding tokens, produce coherent and contentful text \cite{see-etal-2019-massively}, and importantly, capture some commonsense and world knowledge \cite{davison-etal-2019-commonsense, petroni-etal-2019-language}. 

To induce conditional text generation behavior, \citet{Radford2019LanguageMA} propose to add the context tokens (e.g., question and answer) before a special token for the generation start. But for visual-textual tasks, the rationale generation has to be conditioned not only on textual context, but also on an image.

\subsection{Outline of Full-Stack Visual Understanding}
\label{sec:vision_models}

We first outline types of visual information and associated external models that lead to the full-stack visual understanding. Specifications of these models and features that we use appear in Table \ref{tab:vision_models_details}. 

Recognition-level understanding of an image begins with identifying the objects present within it. To this end, we use an object detector that predicts objects present in an image, their labels (e.g., ``cup or ``chair''), bounding boxes, and the boxes' hidden representations (Fig.\ \ref{fig:object_detection}). 

The next step of recognition-level understanding is capturing relations between objects. A computer vision task that aims to describe such relations is situation recognition \cite{yatskar2016}. We use a model for \textit{grounded} situation recognition \cite[Fig.\  \ref{fig:cv_tools_gsr};][]{Pratt2020GroundedSR} that predicts the most prominent activity in an image (e.g., ``surfing''), roles of entities engaged in the activity (e.g., ``agent'' or ``tool''), the roles' bounding boxes, and the boxes' hidden representations. 

The object detector and situation recognizer focus on recognition-level understanding. But visual understanding also requires attributing mental states such as beliefs, intents, and desires to people participating in an image. In order to achieve this, we use the output of \visualcomet \cite[Fig.\ \ref{fig:cv_tools_vcg};][]{Park2020VisualCG}, another \gpt-based model that generates commonsense inferences, i.e.\ events before and after as well as people’s intents, given an image and a description of an event present in the image. 

\subsection{Fusion of Visual and Textual Input}
\label{sec:model_input}
We now describe how we format outputs of the external models (\sect{sec:vision_models}; Table \ref{tab:vision_models_details}) to augment \gpt's input with visual information. 

We explore two ways of extending the input. The first approach adds a vision model's \textit{textual} output (e.g., object labels such as ``food'' and ``table'') before the textual context (e.g., question and answer). Since everything is textual, we can directly embed \textit{each} token using the \gpt's embedding layer, i.e., by summing the corresponding token,
segmentation, and position embeddings.\footnote{The segment embeddings are (to the best of our knowledge) first introduced in \citet{devlin-etal-2019-bert} to separate input elements from different sources in addition to the special separator tokens.}
We call this kind of input fusion \textsc{uniform}.

This is the simplest way to extend the input, but it is prone to propagation of errors from external vision models.  
Therefore, we explore using vision models' \textit{embeddings} for regions-of-interest (RoI) in the image that show relevant entities.\footnote{The entire image is also a region-of-interest.} For each RoI, we sum its visual embedding (described later) with the three \gpt's embeddings (token, segment, position) for a special ``unk'' token and pass the result to the following \gpt blocks.\footnote{Visual embeddings and object labels do not have a natural sequential order among each other, so we assign position zero to them.} After all RoI embeddings, each following token (question, answer, rationale, separator tokens) is embedded similarly, by summing the three \gpt's embeddings and a visual embedding of the entire image. 

We train and evaluate our models with different fusion types and visual features separately to analyze where the improvements come from. We provide details of feature extraction in App.\  \sect{sec:appendix_input_details}.

\paragraph{Visual Embeddings} We build visual embeddings from bounding boxes' hidden representations (the feature vector prior to the output layer) and boxes' coordinates (the top-left, bottom-right coordinates, and the fraction of image area covered). We project bounding boxes' feature vectors as well as their coordinate vectors to the size of \gpt embeddings. We sum projected vectors and apply the layer normalization. We take a different approach for  \visualcomet embeddings, since they are not related to regions-of-interest of the input image (see \sect{sec:vision_models}). In this case, as visual embeddings, we use \visualcomet embeddings that signal to start generating \textit{before}, \textit{after}, and \textit{intent} inferences, and since there is no representation of the entire image, we do not add it to the question, answer, rationale, separator tokens.

%% file: tables/3_vision_models.tex
\begin{table*}[t]
\resizebox{\textwidth}{!}{
\begin{tabular}{lm{0.31\textwidth}m{0.345\textwidth}m{0.345\textwidth}}
\toprule
& \textbf{Object Detector} & \textbf{Grounded Situation Recognizer} & \textbf{Visual Commonsense Graphs} \\
\midrule
Understanding & Basic  & Semantics & Pragmatics\\
\midrule
Model name                   & Faster R-CNN \cite{Ren2015FasterRT} & \textsc{JSL} \cite{Pratt2020GroundedSR} & \textsc{VisualComet} \cite{Park2020VisualCG}\\
  \arrayrulecolor{black!20}\midrule
Backbone & ResNet-50  \cite{He2016DeepRL} & RetinaNet \cite{Lin2017FocalLF}, ResNet-50, LSTM & GPT-2 \cite{Radford2019LanguageMA}\\
  \arrayrulecolor{black!20}\midrule
Pretraining data & ImageNet \cite{Deng2009ImageNetAL} & ImageNet, COCO & OpenWebText \cite{Gokaslan2019OpenWeb} \\
  \arrayrulecolor{black!20}\midrule
Finetuning data & COCO \cite{Lin2014MicrosoftCC} & SWiG  \cite{Pratt2020GroundedSR} & VCG \cite{Park2020VisualCG} \\
  \arrayrulecolor{black}\midrule
\uniform & ``non-person`` object labels                 & top activity and its roles                                         & top-5 \textit{before, after, intent} inferences        \\
 \arrayrulecolor{black}\midrule
\hybrid  & Faster R-CNN's object boxes' representations and coordinates & JSL's role boxes' representations and coordinates                                       & \textsc{VisualComet}'s embedding for special tokens that signal the start of \textit{before, after, intent} inference\ \\
 \arrayrulecolor{black}\bottomrule
\end{tabular}}
\caption{Specifications of external vision models and their outputs that we use as features for visual adaptation.}
\label{tab:vision_models_details}
\end{table*}

%% file: tables/4_datasets.tex
\begin{table*}[t]
\resizebox{\textwidth}{!}{
\begin{tabular}{llllll}
\toprule
\textbf{Dataset} & \textbf{Task} & \textbf{Train} & \textbf{Dev} & \textbf{Expected Visual Understanding} \\
\midrule
\vcr \cite{Zellers2018FromRT}  & \makecell[l]{visual commonsense reasoning\\(question answering)}     &       212,923     &  26,534                   & \makecell[l]{higher-order cognitive, commonsense,\\ recognition}  \\
\midrule
\esnlive \cite{Do2020eSNLIVE20CV} & \multirowcell{2}[0pt][l]{visual-textual entailment} & 511,112$^\dagger$ &      17,133$^\dagger$ &  \multirow{2}{*}{higher-order cognitive, recognition} \\ 
$\neg$ \textsc{neutral}  &  &   341,095     &      13,670     &               \\
\midrule
\vqae \cite{Li2018VQAEEE}   &    visual question answering &      181,298    &    88,488         & recognition         \\    
\bottomrule
\end{tabular}
}
\caption{Specifications of the datasets we use for rationale generation. \citet{Do2020eSNLIVE20CV} 
re-annotate the \textsc{snli-ve} dev and test splits due to the high labelling error of the \textit{neutral} class  \cite{vu-etal-2018-grounded}. Given the remaining errors in the training split, we generate rationales only for entailment and contradiction examples. $\dagger$\citet{Do2020eSNLIVE20CV} report 529,527 and 17,547 training and validation examples, but the available data with explanation is smaller. 
}
\label{tab:datasets}
\end{table*}

%% file: 4_experiments.tex
\section{Experiments}
\label{sec:experiments}

For all experiments, we visually adapt and fine-tune the original \gpt with 117M parameters. We train our models using the language modeling loss computed on rationale tokens.\footnote{See Table \ref{tab:gpt_hyperparameters} (\sect{subsec:appendix_training_details}) for hyperparameter specifications.}

\input{figs/4_entailment_artifact}

\paragraph{Tasks and Datasets}
\label{para:datasets}
We consider three tasks and datasets shown in Table~\ref{tab:datasets}.  Models for \vcr and \vqa are given a question about an image, and they predict the answer from a candidate list. Models for visual-textual entailment are given an image (that serves as a premise) and a textual hypothesis, and they predict an entailment label between them. The key difference among the three tasks is the level of required visual understanding. 

We report here the main observations about how the datasets were collected, while details are in the Appendix \sect{sec:appendix_datasets}. Foremost, only \vcr rationales are human-written for a given problem instance. Rationales in \vqae are extracted from image captions relevant for question-answer pairs \cite{Goyal2017MakingTV} using a constituency parse tree. To create a dataset for explaining visual-textual entailment, \esnlive, \citet{Do2020eSNLIVE20CV} combined the \textsc{snli-ve} dataset \cite{Xie2019VisualEA}  for \textit{visual-textual} entailment and the \esnli dataset \cite{camburu_e_snli_2019} for explaining \textit{textual} entailment.

We notice that this methodology introduced a data collection artifact for entailment cases. To illustrate this, consider the example in Figure \ref{fig:e_snli_ve_entailment_problem}. In visual-textual entailment, the premise is the image. Therefore, there is no reason to expect that a model will build a rationale around a word that occurs in the textual premise it has never seen (``jumping''). We will test whether models struggle with entailment cases.

\input{tables/4_visual_plausibility}

\paragraph{Human Evaluation} For evaluating our models, we follow \citet{camburu_e_snli_2019} who show that \textsc{bleu} \cite{papineni-etal-2002-bleu} is not reliable for evaluation of rationale generation, and hence use human evaluation.\footnote{This is based on a low inter-annotator BLEU-score between three human rationales for the same NLI example.} We believe that other automatic sentence similarity measures are also likely not suitable due to a similar reason; multiple rationales could be plausible, although not necessarily paraphrases of each other (e.g., in Figure \ref{fig:generation_examples} both generated and human rationales are plausible, but they are not strict paraphrases).\footnote{In Table \ref{tab:appendix_captioning_measures} (\sect{subsec:appendix_training_details}), we report automatic captioning measures for the best \modelname for each dataset. These results should be used only for reproducibility and not as measures of rationale plausibility.} Future work might consider newly emerging \textit{learned} evaluation measures, such as \textsc{bleurt} \cite{sellam-etal-2020-bleurt}, that could learn to capture non-trivial semantic similarities between sentences beyond surface overlap. 

We use Amazon Mechanical Turk to crowdsource human judgments of generated rationales according to different criteria. Our instructions are provided in the Appendix \sect{appendix:crowdsourcing}. For \vcr, we randomly sample one QA pair for each movie in the development split of the dataset, resulting in $244$ examples for human evaluation. For \vqa and \esnlive, we randomly sample $250$ examples from their development splits.\footnote{The size of evaluation sample is a general problem of generation evaluation, since human evaluation is crucial but expensive. Still, we evaluate $\sim$2.5 more instances per each of 24 dataset-model combinations than related work \cite{camburu_e_snli_2019, Do2020eSNLIVE20CV, Narang2020WT5TT}; and each instance is judged by 3 workers.} We did not use any of these samples to tune any of our hyperparameters. Each generation was evaluated by $3$ crowdworkers. The workers were paid $\sim$\$13 per hour.

\paragraph{Baselines}

The main objectives of our evaluation are to assess whether (i) proposed visual features help \gpt generate rationales that support a given answer or entailment label better (\textbf{visual plausibility}), and whether (ii) models that generate more plausible rationales are less likely to mention content that is irrelevant to a given image (\textbf{visual fidelity}). As a result, a text-only \gpt approach represents a meaningful baseline to compare to.

In light of work exposing predictive data artifacts \cite[e.g.,][]{gururangan-etal-2018-annotation},
we estimate the effect of artifacts by reporting the difference between visual plausibility of the text-only baseline and plausibility of its rationales assessed without looking at the image (\textbf{textual plausibility}). If both are high, then there are problematic lexical cues in the datasets. Finally, we report estimated \textbf{plausibility of human rationales} to gauge what has been solved and what is next.\footnote{Plausibility of human-written rationales is estimated from our evaluation samples.}

\subsection{Visual Plausibility}
\label{sec:plausibility_results}
We ask workers to judge whether a rationale supports a given answer or entailment label 
in the context of the image (\textit{visual plausibility}). They could select a label from $\{$\textit{yes, weak yes, weak no, no}$\}$. We later merge \textit{weak yes} and \textit{weak no} to \textit{yes} and \textit{no}, respectively. We then calculate the ratio of \textit{yes} labels for each rationale and report the average ratio in a sample.\footnote{We follow the related work \cite{camburu_e_snli_2019, Do2020eSNLIVE20CV, Narang2020WT5TT} in using yes/no judgments. We introduced weak labels because they help evaluating cases with a slight deviation from a clear-cut judgment.}

We compare the text-only \gpt with visual adaptations in Table \ref{tab:results_plausibility_with_image}. We observe that {\small\textsc{GPT-2}}'s visual plausibility benefits from some form of visual adaptation for all tasks. The improvement is most visible for \vqae, followed by \vcr, and then \esnlive (all). We suspect that the minor improvement for \esnlive is caused by the entailment-data artifact. Thus, we also report the visual plausibility for entailment and contradiction cases separately. The results for contradiction hypotheses follow the trend that is observed for \vcr and \vqae. In contrast, visual adaption does not help rationalization of entailed hypotheses. These findings, together with the fact that we have already discarded neutral hypotheses due to the high error rate, raise concern about the \esnlive dataset. Henceforth, we report entailment and contradiction separately, and focus on contradiction when discussing results. We illustrate rationales produced by \modelname in Figure \ref{fig:generation_examples}, and provide additional analyses in the Appendix \sect{sec:appendix_results}.

\subsection{Effect of Visual Features}

We motivate different visual features with varying levels of visual understanding (\sect{sec:vision_models}).  We reflect on our assumptions about them in light of the visual plausibility results in Table \ref{tab:results_plausibility_with_image}. We observe that \visualcomet, designed to help attribute mental states, indeed results in the most plausible rationales for reasoning in \vcr, which requires a high-order cognitive and commonsense understanding. We propose situation frames to understand relations between objects which in turn can result in better recognition-level understanding. Our results show that situation frames are the second best option for \vcr and the best for \vqa, which supports our hypothesis. The best option for \esnlive (contradiction) is \hybrid fusion of objects, although \uniform situation fusion is comparable. Moreover, \visualcomet is less helpful for \esnlive compared to objects and situation frames. This suggests that visual-textual entailment in \esnlive is perhaps focused on recognition-level understanding more than it is anticipated. 

One fusion type does not dominate across datasets (see an overview in Table \ref{tab:hybrid_vs_uniform} in the Appendix  \sect{sec:appendix_results}). We hypothesize that the source domain of the pretraining dataset of vision models as well as their precision can influence which type of fusion works better. A similar point was recently raised by \citet{Singh2020AreWP}. Future work might consider carefully combining both fusion types and multiple visual features. 

\input{tables/4_textual_plausibility}
\input{figs/4_generations}
\input{figs/4_fidelity_vs_plausibility_change}
\input{tables/4_correlation_visual_fidelity_and_plausibility}

\subsection{Textual Plausibility} It has been shown that powerful pretrained LMs can reason about textual input well in the current benchmarks \cite[e.g.,][]{zellers-etal-2019-hellaswag, Khashabi2020UnifiedQACF}. In our case, that would be illustrated with a high plausibility of generated rationales in an evaluation setting where workers are instructed to ignore images (\textit{textual plausibility}).

We report textual plausibility in Table \ref{tab:results_plausibility_no_image}. Text-only \gpt achieves high textual plausibility (relative to the human estimate) for all tasks (except the entailment part of \esnlive), demonstrating good reasoning capabilities of \gpt, when the context image is ignored for plausibility assessment. This result also verifies our hypothesis that generating a textually plausible rationale is easier for models than producing a visually plausible rationale. 
For example, \gpt can likely produce many statements that contradict ``the woman is texting'' (see Figure \ref{fig:generation_examples}), but producing a visually plausible rationale requires conquering another challenge: capturing what is present in the image. 

If both textual and visual plausibility of the text-only \gpt were high, that would indicate there are some lexical cues in the datasets that allow models to ignore the context image. 
The decrease in plausibility performance once the image is shown (cf.\ Tables \ref{tab:results_plausibility_with_image} and  \ref{tab:results_plausibility_no_image}) confirms that the text-only baseline is not able to generate visually plausible rationales by fitting lexical cues.

We notice another interesting result: textual plausibility of visually adapted models is higher than textual plausibility of the text-only \gpt. The following three insights together suggest why this could be the case: (i) the gap between textual plausibility of generated and human rationales shows that generating textual plausible rationales is not solved, (ii) visual models produce rationales that are more visually plausible than the text-only baseline, and (iii) visually plausible rationales are usually textually plausible (see examples in Figure \ref{fig:generation_examples}).

\subsection{Plausibility of Human Rationales}
The best performing models for \vcr and \esnlive (contradiction) are still notably behind the estimated visual plausibility of human-written rationales (see Table \ref{tab:results_plausibility_with_image}). Moreover, plausibility of human rationales is similar when evaluated in the context of the image (visual plausibility) and without the image (text plausibility) because (i) data annotators produce visually plausible rationales since they have accurate visual understanding, and (ii) visually plausible rationales are usually textually plausible. These results show that generating visually plausible rationales for \vcr and \esnlive is still challenging even for our best models.

In contrast, we seem to be closing the gap for \vqae. In addition, due in part to the automatic extraction of rationales, the human rationales in \vqae suffer from a notably lower estimate of plausibility.

\subsection{Visual Fidelity} 
\label{para:hallucination_intro}

We investigate further whether visual plausibility improvements come from better visual understanding. 
We ask workers to judge if the rationale mentions content unrelated to the image, i.e., anything that is not directly visible and is unlikely to be present in the scene in the image. They could select a label from $\{$\textit{yes, weak yes, weak no, no}$\}$. We later merge \textit{weak yes} and \textit{weak no} to \textit{yes} and \textit{no}, respectively. We then calculate the ratio of \textit{no} labels for each rationale. The final \textbf{fidelity} score is the average ratio in a sample.\footnote{We also study assessing fidelity from phrases that are extracted from a rationale (see Appendix \ref{sec:appendix_results}).} 

Figure \ref{fig:hallucination_plausibility} illustrates the relation between visual fidelity and plausibility. For each dataset (except the entailment part of \esnlive), we observe that visual plausibility is larger as visual fidelity increases. We verify this with Pearson's $r$ and show moderate linear correlation in Table \ref{tab:correlation_hallucination_plausibility}. This shows that models that generate more visually plausible rationales are less likely to mention content that is irrelevant to a given image.

%% file: figs/4_entailment_artifact.tex
\begin{figure}[t]
\centering
\includegraphics[width=0.7\columnwidth]{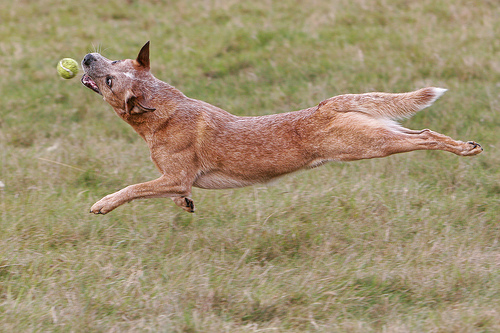}\\
\begin{tabular}{p{0.9\columnwidth}}
\arrayrulecolor{black!0}\toprule
\textbf{Hypothesis}: A dog plays with a tennis ball.\\
\textbf{Label}: Entailment.\\
\textbf{Rationale}: A dog \textit{jumping} is how he plays.\\
\arrayrulecolor{black!30}\midrule
\textbf{Textual premise}: A brown dog is \textit{jumping} after a tennis ball.
\end{tabular}
\caption{An illustrative example of the entailment artifact in \esnlive.}
\label{fig:e_snli_ve_entailment_problem}
\end{figure}

%% file: tables/4_visual_plausibility.tex
\begin{table*}[!ht]
\centering
\resizebox{0.9\textwidth}{!}{
\setlength{\tabcolsep}{8pt}
\centering
\begin{tabular}{l@{\hskip 2mm}l@{\hskip 4mm}l@{\hskip 4mm}lccccc}
\toprule
    &&&&\multirow{2}{*}{\textbf{\vcr}}& \multicolumn{3}{c}{\textbf{\esnlive}} & \multirow{2}{*}{\textbf{\vqae}}\\
\cmidrule{6-8}
    &&&   &  & Contradiction & Entailment & All &  \\
\midrule
    &&& Baseline   & 53.14   & 46.85  & \textbf{46.76} & 46.80 & 47.20 \\
\midrule
\multirow{6}{*}{\rotatebox{90}{\textsc{Rationale}$^{\textsc{VT}}$}} & \multirow{6}{*}{\rotatebox{90}{\textsc{Transformers}}} & \multirow{3}{*}{\rotatebox{90}{\small\uniform}} & Object labels & 54.92     & 58.56 & 36.45  & 46.27  & 54.40 \\
   && & Situation frames      & 56.97       & 59.16 & 38.13 & \textbf{47.47} & 50.93 \\
   && & \viscomet text inferences    & \textbf{60.93}      & 53.75 & 29.26 & 40.13  & 53.47 \\
\cmidrule{3-9}
&&\multirow{3}{*}{\rotatebox{90}{\small\hybrid}}  & Object regions  & 47.40  & \textbf{60.96} & 34.05  & 46.00 & 59.07\\
                      &&   & Situation roles regions   & 47.95 & 51.95 & 37.65 &  44.00 & \textbf{63.33}    \\
                  &&       & \viscomet embeddings & 59.84 & 48.95 & 32.13 & 39.60   & 54.93\\
\midrule
\midrule
&&& Human (estimate)   & 87.16   & 80.78  & 76.98  &  78.67 & 66.53\\
\bottomrule
\end{tabular}
}
\caption{Visual plausibility of random samples of generated and human (gold) rationales. Our baseline is text-only \gpt. The best model is boldfaced.}
\label{tab:results_plausibility_with_image}
\end{table*}

%% file: tables/4_textual_plausibility.tex
\begin{table}[t]
\centering
\resizebox{\columnwidth}{!}{
\begin{tabular}{llcccc}
\toprule
&   & \textbf{\vcr} & \makecell[c]{\textbf{\esnlive}\\\textbf{(contrad.)}} & \makecell[c]{\textbf{\esnlive}\\\textbf{(entail.)}}  & \textbf{\vqae} \\
\midrule
& Baseline   & 70.63 & 74.47   &   \textbf{46.28}   & 68.27 \\
\midrule
\multirow{3}{*}{\rotatebox{90}{\small\uniform}} & Object labels & 75.14 & \textbf{77.48} & 37.17      & 64.80 \\
    & Situation frames & 74.18 & 72.37  &  38.61   & 61.73 \\
    & \viscomet text    & 73.91 & 71.17  &   32.37   & 69.73 \\
\midrule
\multirow{3}{*}{\rotatebox{90}{\small\hybrid}}  & Object regions  & 69.26 & 69.97   & 33.81     & 68.53 \\
                         & Situation roles regions      & 69.81 & 62.46  & 38.61 & 69.47 \\
                         & \viscomet embd.\ & \textbf{81.15} &74.77  & 32.37   & \textbf{76.53}  \\
\midrule
\midrule
& Human (estimate)   & 90.71 &79.58 &   74.34 & 64.27 \\
\bottomrule
\end{tabular}
}
\caption{Plausibility of random samples of human (gold) and generated rationales assessed \textbf{\underline{without}} looking at the image (\textit{textual plausibility}). The best model is boldfaced.}
\label{tab:results_plausibility_no_image}
\end{table}

%% file: figs/4_generations.tex
\begin{figure*}[t]
\centering
\includegraphics[width=\textwidth]{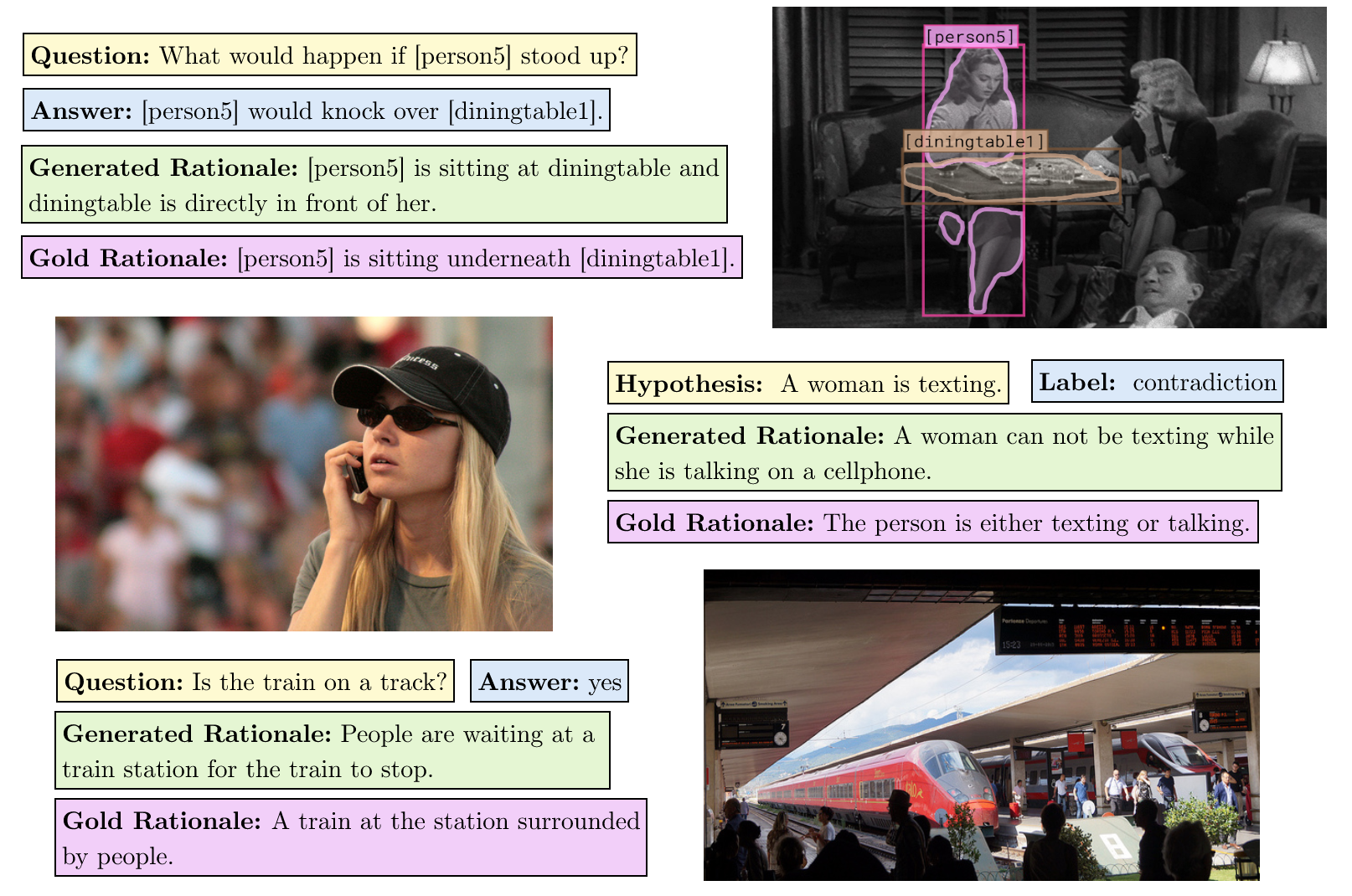}
  \caption{\modelname generations for \vcr (top), \esnlive (contradiction; middle), and \vqae (bottom). We use the best model variant for each dataset (according to results in Table \ref{tab:results_plausibility_with_image}).}
  \label{fig:generation_examples}
\end{figure*}

%% file: figs/4_fidelity_vs_plausibility_change.tex
\begin{figure*}[t]
\centering
\includegraphics[width=\textwidth]{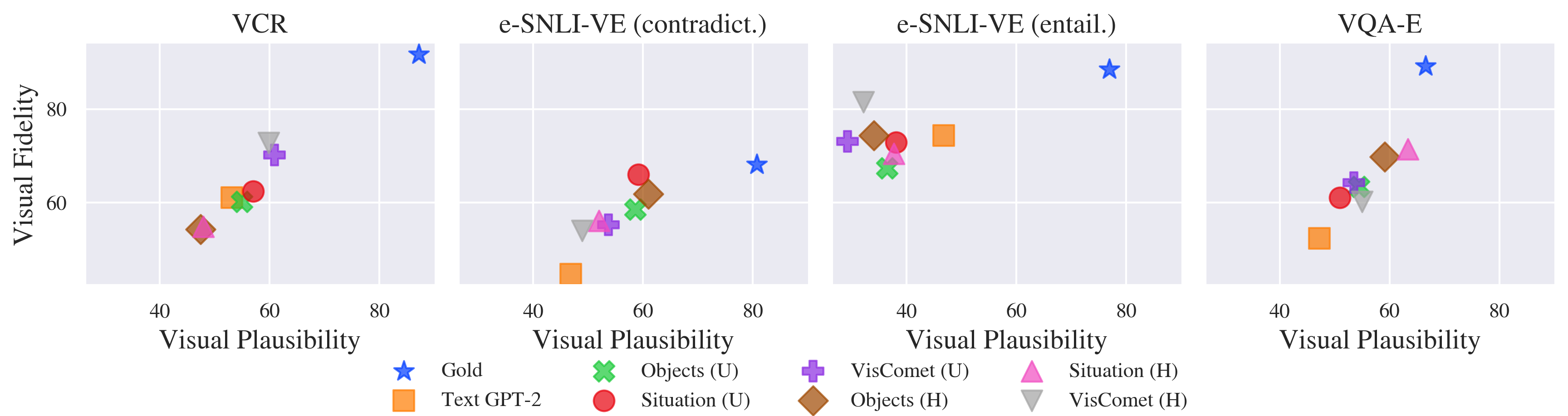}
\caption{The relation between 
visual plausibility (\sect{sec:plausibility_results}) and visual fidelity (\sect{para:hallucination_intro}). We denote \uniform fusion with (U) and \hybrid fusion with (H).}
\label{fig:hallucination_plausibility}
\end{figure*}

%% file: tables/4_correlation_visual_fidelity_and_plausibility.tex
\begin{table*}[t]
\resizebox{\textwidth}{!}{
\begin{tabular}{l@{\hskip 2mm}l@{\hskip 4mm}l@{\hskip 4mm}lcccccccccccc}
\toprule
& && & \multicolumn{3}{c}{\textbf{\vcr}} & \multicolumn{3}{c}{\makecell[c]{\textbf{\esnlive}\\\textbf{(contradict.)}}} & \multicolumn{3}{c}{\makecell[c]{\textbf{\esnlive}\\\textbf{(entail.)}}} &
 \multicolumn{3}{c}{\textbf{\vqae}} \\
 \cmidrule(lr){5-7}\cmidrule(lr){8-10}\cmidrule(lr){11-13}\cmidrule(lr){14-16} 
 &&& & Plaus.\  & Fidelity\     & $r$     & Plaus.\                          & Fidelity\  & $r$                     & Plaus.\    & Fidelity\                 & $r$         & Plaus.\                  & Fidelity\    & $r$              \\
\midrule
&&& Baseline &  53.14 & 61.07 & 0.68 & 46.85 & 44.74 & 0.53 & \textbf{46.76} & \underline{74.34} & 0.50 & 47.20 & 52.40 & 0.61 \\
\midrule
\multirow{6}{*}{\rotatebox{90}{\textsc{Rationale}$^{\textsc{VT}}$}} & \multirow{6}{*}{\rotatebox{90}{\textsc{Transformers}}} & \multirow{3}{*}{\rotatebox{90}{\small\uniform}} & Object labels & 54.92 & 60.25 & 0.73 & 58.56 & 58.56 & 0.55 & 36.45 & 67.39 & 0.58 & 54.40 & 63.47 & 0.54 \\
 & && Situation frames & 56.97 & 62.43 & 0.78 & \underline{59.16} & \textbf{66.07} & 0.37 & \underline{38.13} & 72.90 & 0.51 & 50.93 & 61.07 & 0.53 \\
 &&& \viscomet  text & \textbf{60.93} & \underline{70.22} & 0.62 & 53.75 & 55.26 & 0.45 & 29.26 & 73.14 & 0.49 & 53.47 & 64.27 & 0.66 \\
\cmidrule(lr){3-16}
 &&\multirow{3}{*}{\rotatebox{90}{\small\hybrid}}  & Object regions  & 47.40 & 54.37 & 0.67 & \textbf{60.96} & \underline{61.86} & 0.40 & 34.05 & \underline{74.34} & 0.31 & \underline{59.07} & \underline{69.87} & 0.53 \\
 && & Situ.\ roles regions  & 47.95 & 54.92 & 0.66 & 51.95 & 56.16 & 0.45 & 37.65 & 70.50 & 0.59 & \textbf{63.33} & \textbf{71.47} & 0.62 \\
 & && \viscomet embd.\  & \underline{59.84} & \textbf{72.81} & 0.72 & 48.95 & 54.05 & 0.48 & 32.13 & \textbf{81.53} & 0.41 & 54.93 & 60.27 & 0.59 \\
\midrule
\midrule
 &&& Human (estimate) & 87.16 & 91.67 & 0.58 & 80.78 & 68.17 & 0.28 & 76.98 & 88.49 & 0.43 & 66.53 & 89.20 & 0.35 \\
\bottomrule       
\end{tabular}}
\caption{Visual plausibility (Table \ref{tab:results_plausibility_with_image}; \sect{sec:plausibility_results}), visual fidelity (\sect{para:hallucination_intro}), and Pearson's $r$  that measures linear correlation between the visual plausibility and fidelity. Our baseline is text-only \gpt. The best model is boldfaced and the second best underlined.}
\label{tab:correlation_hallucination_plausibility}
\end{table*}

%% file: 5_related_work.tex
\section{Related Work} 

\paragraph{Rationale Generation}  Applications of rationale generation (see \sect{sec:intro}) can be categorized as text-only, vision-only, or visual-textual. 
Our work belongs to the final category, where we are the first to try to generate rationales for \vcr \cite{Zellers2018FromRT}. The bottom-up top-down attention (BUTD) model \cite{Anderson2018BottomUpAT} has been proposed to incorporate rationales with visual features for \vqae and \esnlive \cite{Li2018VQAEEE, Do2020eSNLIVE20CV}. Compared to BUTD, we use a pretrained decoder and propose a wider range of visual features to tackle comprehensive image understanding.

\paragraph{Conditional Text Generation}
Pretrained LMs have played a pivotal role in open-text generation and conditional text generation. For the latter, some studies trained a LM from scratch conditioned on metadata \citep{zellers_grover_2019} or desired attributes of text \cite{Keskar2019CTRLAC}, while some fine-tuned an already pretrained LM on commonsense knowledge \cite{Bhagavatula2019AbductiveCR} or text attributes \cite{Ziegler2019FineTuningLM}. Our work belongs to the latter group with focus on conditioning on comprehensive image understanding.

\paragraph{Visual-Textual Language Models}
There is a surge of work that proposes visual-textual  pretraining of LMs by predicting masked image regions and tokens \cite[to name a few]{Tan2019LXMERTLC, Lu2019ViLBERTPT, Chen2019UNITERLU}. We construct input elements of our models following the \textsc{vl-bert} architecture \cite{Su2019VLBERTPO}. 
Despite their success, these models are not suitable for generation due to pretraining with the masked LM objective. \citet{Zhou2019UnifiedVP} aim to address that, but they pretrain their decoder from scratch using 3M images with weakly-associated captions \cite{sharma-etal-2018-conceptual}. This makes their decoder arguably less powerful compared to LMs that are pretrained with remarkably more (diverse) data such as \gpt. \citet{Ziegler2019EncoderAgnosticAF} augment \gpt with a feature vector for the entire image and evaluate this model on image paragraph captioning. Some work extend pretrained LM to learn video representations from sequences of visual features and words, and show improvements in video captioning \cite{Sun2019ContrastiveBT, Sun2019VideoBERTAJ}. Our work is based on fine-tuning \gpt with features that come from visual object recognition, grounded semantic frames, and visual commonsense graphs. The latter two features have not been explored yet in this line of work.

\section {Discussion and Future Directions}

\paragraph{Rationale Definition} The term \textit{interpretability} is used to refer to multiple concepts. Due to this, criteria for explanation evaluation depend on one's definition of interpretability \cite{mythos-lipton-2016, DoshiVelez2017TowardsAR, Jacovi2020TowardsFI}. In order to avoid problems arising from ambiguity, we reflect on our definition. We follow \citet {Harrison2018RationalizationAN} who define AI \textit{rationalization} as a process of generating rationales of a model's behavior as if a human had performed the behavior. 

\paragraph{Jointly Predicting and Rationalizing} We narrow our focus on improving generation models and assume gold labels for the end-task. Future work can extend our model to an end-to-end \cite{Narang2020WT5TT} or a pipeline model \cite{camburu_e_snli_2019, rajani-etal-2019-explain, Jain2020LearningTF} for producing both predictions and natural language rationales.
We expect that the \textit{explain-then-predict} setting \cite{camburu_e_snli_2019} is especially relevant for rationalization of commonsense reasoning. In this case, relevant information is not in the input, but inferred from it, which makes extractive explanatory methods based on highlighting parts of the input unsuitable. A rationale generation model brings relevant information to the surface, which can be passed to a prediction model. This makes rationales intrinsic to the model, and tells the user what the prediction should be based on. \citet{kumar-talukdar-2020-nile} highlight that this approach resembles post-hoc methods with the label and rationale being produced jointly (the \textit{end-to-end predict-then-explain} setting). Thus, all but the pipeline predict-then-explain approach are suitable extensions of our models. A  promising line of work trains end-to-end models for joint rationalization and prediction from weak supervision \cite{Latcinnik2020ExplainingQA, Shwartz2020UnsupervisedCQ}, i.e., without human-written rationales. 

\paragraph{Limitations} Natural language rationales are easily understood by lay users who consequently feel more convinced and willing to use the model \cite{Miller2019ExplanationIA, Ribera2019CanWD}. Their limitation is that they can be used to persuade users that the model is reliable when it is not \cite{Bansal2020DoesTW}---an ethical issue raised by \citet{Herman2017ThePA}. This relates to the pipeline  predict-then-explain setting, where a predictor model and a post-hoc explainer model are completely independent. However, there are other settings where generated rationales are intrinsic to the model by design (end-to-end predict-then-explain, both end-to-end and pipeline explain-then-predict). As such, generated rationales are more associated with the reasoning process of the model. We recommend that future work develops rationale generation in these settings, and aims for \textit{sufficiently faithful} models as recommended by \citet{Jacovi2020TowardsFI}, \citet{wiegreffe-pinter-2019-attention}.

%% file: 6_conclusions.tex
\section{Conclusions}

We present \modelname, an integration of a pretrained text generator with semantic and pragmatic visual features. %
These features can improve visual plausibility and fidelity of generated rationales for visual commonsense reasoning, visual-textual entailment, and visual question answering. %
This represents progress in tackling important, but still relatively unexplored research direction; rationalization of complex reasoning for which explanatory approaches based solely on highlighting parts of the input are not suitable.

%% file: supplement.tex
\section{Experimental Setup}

\subsection{Deatils of GPT-2}
\label{sec:appendix_gpt_details}
Input to \gpt is text that is split into subtokens\footnote{Also known as wordpieces or subwords.} \cite{sennrich-etal-2016-neural}. Each subtoken embedding is added to a so-called positional embedding that signals the order of the subtokens in the sequence to the transformer blocks. The \gpt's pretraining corpus is 
OpenWebText corpus \cite{Gokaslan2019OpenWeb} which consists of 8 million Web documents extracted from URLs shared on Reddit. Pretraining on this corpus has caused degenerate and biased behaviour of \gpt \cite[among others]{sheng-etal-2019-woman, wallace-etal-2019-universal, Gehman2020RealToxicityPromptsEN}. Our models likely have the same issues since they are built on \gpt.

\subsection{Details of Datasets with Human Rationales}
\label{sec:appendix_datasets}

We obtain the data from the following links: 
\begin{compactitem}
\item \url{https://visualcommonsense.com/download/}
\item \url{https://github.com/virginie-do/e-SNLI-VE}
\item \url{https://github.com/liqing-ustc/VQA-E}
\end{compactitem}

Answers in \vcr are full sentences, and in \vqa single words or short phrases. All annotations in \vcr are authored by crowdworkers in a single data collection phase. Rationales in \vqae are extracted from relevant image captions for question-answer pairs in \vqa \textsc{\small v2} \cite{Goyal2017MakingTV} using a constituency parse tree. The overall quality of \vqae rationales is 4.23/5.0 from human perspective.

The \esnlive dataset is constructed from a series of additions and changes of the \textsc{snli} dataset for \textit{textual} entailment \cite{bowman-etal-2015-large}. The \textsc{snli} dataset  is collected by using captions in Flickr30k \cite{young-etal-2014-image} as textual premises and crowdsourcing  hypotheses.\footnote{Captions tend to be literal scene descriptions.} The \esnli dataset \cite{camburu_e_snli_2019} adds crowdsourced explanations to \textsc{snli}. The \textsc{snli-ve} dataset \cite{Xie2019VisualEA} for \textit{visual-textual} entailment is constructed from \textsc{snli} by replacing textual premises with corresponding Flickr30k images. Finally, \citet{Do2020eSNLIVE20CV} combine \textsc{snli-ve} and \esnli to produce a dataset for explaining \textit{visual-textual} entailment. They re-annotate the dev and test splits due to the high labelling error of the \textit{neutral} class in \textsc{snli-ve} that is reported by \citet{vu-etal-2018-grounded}. 

\input{tables/appendix_image_sources}

\subsection{Details of External Vision Models}
\label{sec:appendix_vision_models_details}
In Table \ref{tab:appendix_image_sources}, we report sources of images that were used to train external vision models and images in the end-task datasets. 

\subsection{Details of Input Elements}
\label{sec:appendix_input_details}

\paragraph{Object Detector} For \uniform fusion, we use labels for objects other that people because \textit{person} label occurs in every example for \vcr. We use only a single instance of a certain object label, because repeating the same label does not give new information to the model. The maximum number of subtokens for merged object labels is determined from merging all object labels, tokenizing them to subtokens, and set the maximum to the length at the ninety-ninth percentile calculated from the \vcr training set. For \hybrid fusion, we use hidden representation of all objects because they differ for different detections of objects with the same label. These representations come from the feature vector prior to the output layer of the detection model. The maximum number of objects is set to the object number at the 99th percentile calculated from the \vcr training set.  

\paragraph{Situation Recognizer} For \uniform fusion, we consider only the best verb because the top verbs are often semantically similar (e.g.\ \textit{eating} and \textit{dining}; see Figure 13 in \citet{Pratt2020GroundedSR} for more examples). We define a structured format for the output of a situation recognizer. For example, the situation predicted from the first image in Figure \ref{fig:generation_examples}, is assigned the following structure ''\texttt{\small  <|b\char`_situ|> <|b\char`_verb|> dining <|e\char`_verb|> <|b\char`_agent|> people <|e\char`_agent|> <|b\char`_place|> restaurant <|e\char`_place|> <|e\char`_situ|>}''. We set the maximum situation length to the length at the ninety-ninth percentile calculated from the \vcr training set. 

 \paragraph{\textsc{VisualComet}} The input to \visualcomet is an image, question, and answer for \vcr and \vqae; only image for \esnlive. Unlike  situation frames, top-k \visualcomet inferences are diverse. We merge top-5 before, after, and intent inferences. We calculate the length of merged inferences in number of subtokens and set the maximum \visualcomet length to the length at the ninety-ninth percentile calculated from the \vcr training set. 
 
\subsection{Training Details}
\label{subsec:appendix_training_details}

We use the original \gpt version with 117M parameters. It consists of 12 layers, 12 heads for each layer, and the size of a model dimension set to 768. We report other hyperaparametes in Table \ref{tab:gpt_hyperparameters}. All of them are manually chosen due to the reliance on human evaluation. 
In Table \ref{tab:appendix_captioning_measures}, for reproducibility, we report captioning measures of the best \modelname variants. Our implementation uses
the HuggingFace \texttt{transformers} library \cite{wolf2019huggingfaces}.\footnote{\url{https://github.com/huggingface/transformers}}

\subsection{Crowdsourcing Human Evaluation}
\label{appendix:crowdsourcing}
We perform human evaluation of the generated rationales through crowdsourcing on the Amazon Mechanical Turk platform. Here, we provide the full set of \textbf{Guidelines} provided to workers:

\begin{itemize}
  \setlength\itemsep{0em}
    \item First, you will be shown a (i) Question, (ii) an Answer (presumed-correct), and (iii) a Rationale. You'll have to judge if the rationale supports the answer. 
    \item Next, you will be shown the same question, answer, rationale, and an associated image. You'll have to judge if the rationale supports the answer, in the context of the given image. 
    \item You'll judge the grammaticality of the rationale. Please ignore the absence of periods, punctuation and case. 
    \item Next, you'll have to judge if the rationale mentions persons, objects, locations or actions unrelated to the image---i.e. things that are not directly visible and are unlikely to be present to the scene in the image. 
    \item Finally, you'll pick the NOUNS, NOUN PHRASES and VERBS from the rationale that are unrelated to the image. 
\end{itemize}

We also provide the following additional \textbf{tips}:
\begin{itemize}
\setlength\itemsep{0em}
    \item Please ignore minor grammatical errors---e.g. case sensitivity, missing periods etc. 
    \item Please ignore gender mismatch---e.g. if the image shows a male, but the rationale mentions female. 
    \item Please ignore inconsistencies between person and object detections in the QUESTION / ANSWER and those in the image---e.g. if a pile of papers is labeled as a laptop in the image. Do not ignore such inconsistencies for the rationale. 
    \item When judging the rationale, think about whether it is plausible. 
    \item If the rationale just repeats an answer, it is not considered as a valid justification for the answer.
\end{itemize}

\section{Additional Results}
\label{sec:appendix_results}

We provide the following additional results that complement the discussion in Section \ref{sec:experiments}:
\begin{compactitem}
\item a comparison between \uniform and \hybrid fusion in Table \ref{tab:hybrid_vs_uniform},
\item an investigation of fine-grained visual fidelity in Table \ref{tab:appendix_fidelity_breakdown},
\item additional analysis of \modelname to support future developments.
\end{compactitem}

\input{tables/appendix_HPs}

\input{tables/appendix_captioning_measures}

\input{tables/appendix_hybrid_vs_uniform_fusion}
\input{tables/4_grammaticality}
\input{tables/appendix_fidelity_breakdown}

\paragraph{Fine-Grained Visual Fidelity}  At the time of running human evaluation, we did not know whether judging visual fidelity is a hard task for workers. To help them focus on relevant parts of a given rationale and to make their judgments more comparable, we give workers a list of nouns, noun phrases, as well as verb phrases with negation, without adjuncts. We ask them to pick phrases that are unrelated to the image. For each rationale, we calculate the ratio of nouns that are relevant over the number of all nouns. We call this \textbf{``entity fidelity''} because extracted nouns are mostly concrete (opposed to abstract). Similarly, from noun phrases judgments, we calculate \textbf{``entity detail fidelity''}, and from verb phrases \textbf{``action fidelity''}. Results in Table \ref{tab:appendix_fidelity_breakdown} show close relation between the overall fidelity judgment and entity fidelity. Furthermore, for the case where the top two models have close fidelity (\visualcomet models for \vcr), the fine-grained analysis shows where the difference comes from (in this case from action fidelity). Despite possible advantages of fine-grained fidelity, we observe that is less correlated with plausibility compared to the overall fidelity. 

\paragraph{Additional Analysis} We ask workers to judge grammatically of rationales. We instruct them to ignore some mistakes such as absence of periods and mismatched gender (see \sect{appendix:crowdsourcing}). Table \ref{tab:grammaticality} shows that the ratio of grammatical rationales is high for all model variants. 

We measure similarity of generated and gold rationales to question (hypothesis) and answer. Results in Tables \ref{tab:appendix_similarities_generated_rationale}--\ref{tab:appendix_similarities_gold_rationale} show that generated rationales repeat the question (hypothesis) more than human rationales. We also observe that gold rationales in \esnlive are notably more repetitive than human rationales in other datasets. 

In Figure \ref{fig:appendix_analysis_input_lenght}, we show that the length of generated rationales is similar for plausible and implausible rationales, with the exception of \esnlive for which implausible rationales tend to be longer than plausible. We show that plausible rationales tend to rationalize slightly shorter textual context in \vcr (question and answer) and \esnlive (hypothesis).

Finally, in Figure \ref{fig:appendix_plausibility_variation}, we show that there is more variation across $\{$\textit{yes, weak yes, weak no, no}$\}$ labels for our models than for human rationales.  

In summary, future developments should improve generations such that they repeat textual context less, handle long textual contexts, and produce generations that humans will find more plausible with high certainty.

\input{tables/appendix_similarites_between_q_a_and_r}
\clearpage
\input{figs/appendix_analysis}

%% file: tables/appendix_image_sources.tex
\begin{table*}[t]
\centering
\resizebox{0.55\textwidth}{!}{
\begin{tabular}{ll}
\toprule
\textbf{Dataset}   & \textbf{Image Source}             \\
\midrule
COCO      & Flickr                   \\
\esnlive & Flickr \cite[SNLI;][]{bowman-etal-2015-large}          \\
ImageNet   & different search engines \\
SWiG      & Google Search \cite[imSitu;][]{yatskar2016}   \\
VCG, \vcr  & movie clips \cite{Rohrbach2016MovieD}, Fandango$^\dagger$             \\
\vqae    & Flickr (COCO)           \\
\bottomrule
\end{tabular}}
\caption{Image sources. $\dagger$ \url{https://www.youtube.com/user/movieclips}}
\label{tab:appendix_image_sources}
\end{table*}

%% file: tables/appendix_HPs.tex
\begin{table*}[t!]
    \centering
    \small
    \begin{tabular}{cc}
      \toprule
      \textbf{Computing Infrastructure} & Quadro RTX 8000 GPU \\
      \midrule
      \textbf{Model implementation} & \url{https://github.com/allenai/visual-reasoning-rationalization}\\
      \bottomrule
    \end{tabular}
    
    \vspace{3mm}\begin{tabular}{ll}
        \toprule
        \textbf{Hyperparameter} & \textbf{Assignment}  \\
        \midrule
        number of epochs & 5 \\
        \midrule
        batch size & 32 \\
        \midrule
        learning rate & 5e-5\\
          \midrule
        max question length & 19 \\
        \midrule 
        max answer length & 23 \\
        \midrule 
        max rationale length & 50 \\
        \midrule 
        max merged object labels length & 30 \\
        \midrule
        max situation's structured description length & 17\\
        \midrule 
        max \visualcomet merged text inferences length & 148\\
        \midrule 
        max input length & 93, 98, 123, 102, 112, 241 \\
        \midrule 
        max objects embeddings number & 28\\
        \midrule
        max situation role embeddings number & 7\\
        \midrule 
        dimension of object and situation role embeddings & 2048 \\
        \midrule
        decoding & greedy\\
        \bottomrule
    \end{tabular}
    \caption{Hyperparameters for \modelname. The length is calculated in number of subtokens including special separator tokens for a given input type (e.g., begin and end separator tokens for a question). We calculate the maximum input length by summing the maximum lengths of input elements for each model separately. A training epoch for models with shorter maximum input length $\sim$30 minutes and for the model with the longest input $\sim$2H.} 
    \label{tab:gpt_hyperparameters}
\end{table*}

%% file: tables/appendix_captioning_measures.tex
\begin{table*}[t]
\centering
\resizebox{\textwidth}{!}{
\begin{tabular}{lcccc}
\toprule
 & \textbf{\vcr} & \makecell[c]{\textbf{\esnlive}\\\textbf{(contradict.)}} & \makecell[c]{\textbf{\esnlive}\\\textbf{(entail.)}} & \textbf{\vqae} \\
 & \visualcomet \uniform & Situation Frame \uniform & Text-Only \gpt & Situation Frame \hybrid \\
 \midrule
BLEU-1 & 20.98 & 32.18 & 33.09 & 36.64 \\
BLEU-2 & 12.15 & 20.35 & 22.55 & 22.48 \\
BLEU-3 & 7.52 & 13.90 & 15.78 & 14.33 \\
BLEU-4 & 4.98 & 9.50 & 11.37 & 9.47 \\
METEOR & 12.21 & 19.29 & 20.09 & 19.33 \\
ROUGE-L & 23.08 & 27.25 & 27.74 & 35.31 \\
CIDEr & 37.22 & 71.37 & 73.35 & 94.89 \\
\bottomrule
\end{tabular}
}
\caption{We report standard automatic captioning measure for the best \modelname for each dataset (according to results in Table \ref{tab:results_plausibility_with_image}; \sect{sec:plausibility_results}), except for \esnlive for which we use \uniform fusion of situation frames instead of object labels, because they have comparable plausibility, but situation frames result in better fidelity. We use the \textbf{\underline{entire development sets}} for this evaluation.}
\label{tab:appendix_captioning_measures}
\end{table*}

%% file: tables/appendix_hybrid_vs_uniform_fusion.tex
\begin{table*}[t]
\centering
\resizebox{0.5\textwidth}{!}{
\begin{tabular}{llcc}
\toprule
 &  & \textbf{\uniform} & \textbf{\hybrid} \\
\midrule
\multirow{3}{*}{\vcr} & Objects & 7.51 & - \\
 & Situation frame & 9.02 & - \\
 & \visualcomet  & 1.09 & - \\
\midrule
 \multirowcell{3}[0pt][l]{\esnlive\\(contradiction)} & Objects & - & 2.40 \\
 & Situation frame & 7.21 & - \\
 & \visualcomet & 4.80 & - \\
\midrule
 \multirowcell{3}[0pt][l]{\esnlive\\(entailment)} & Objects & 2.40 & - \\
 & Situation frame & 0.48 & - \\
 & \visualcomet & - & 2.88 \\
\midrule
\multirow{3}{*}{\vqae} & Objects & - & 4.67 \\
 & Situation frame & - & 12.40 \\
 & \visualcomet & -  & 1.47\\
\bottomrule
\end{tabular}
}
\caption{Comparison of \hybrid and \uniform fusion visual plausibility results that are reported in Table \ref{tab:results_plausibility_with_image} (\sect{sec:plausibility_results}). The number shows the difference in visual plausibility between the fusion type in a given column and the other column. The number is placed in the column with better fusion type for a given task and feature.}
\label{tab:hybrid_vs_uniform}
\end{table*}

%% file: tables/4_grammaticality.tex
\begin{table*}[t]
\centering
\resizebox{0.7\textwidth}{!}{
\centering
\begin{tabular}{l@{\hskip 2mm}l@{\hskip 4mm}l@{\hskip 4mm}lcccc}
\toprule
    &&&   & \textbf{\vcr} & \makecell[c]{\textbf{\esnlive}\\\textbf{(contradict.)}} & \makecell[c]{\textbf{\esnlive}\\\textbf{(entail.)}} & \textbf{\vqae} \\
\midrule
 & &  & Baseline & 92.49 & 94.29 & \underline{86.81} & 96.53 \\
 \midrule
\multirow{6}{*}{\rotatebox{90}{\textsc{Rationale}$^{\textsc{VT}}$}} & \multirow{6}{*}{\rotatebox{90}{\textsc{Transformers}}} & \multirow{3}{*}{\rotatebox{90}{\small\uniform}} & Object labels & 92.62 & \textbf{96.10} & \textbf{87.05} & 97.20 \\
 &  &  & Situation frames & 92.62 & 94.89 & 86.33 & 95.07 \\
 &  &  & \viscomet text inferences & \underline{94.54} & 94.89 & 82.97 & 97.73 \\
 \cmidrule{3-8}
&&\multirow{3}{*}{\rotatebox{90}{\small\hybrid}} & Object regions & 93.03 & \underline{95.50} & 84.65 & 96.67 \\
 &  &  & Situation roles regions & 90.03 & 94.59 & 86.33 & \underline{96.67} \\
 &  &  & \viscomet embeddings & \textbf{96.31} & 95.20 & 84.65 & \textbf{98.13} \\
 \midrule
 \midrule

 &  &  &Human (estimate) & 95.22 & 87.69 & 86.33 & 94.67\\
 \bottomrule
\end{tabular}
}
\caption{The ratio of grammatically correct rationales (according to human evaluation) in random samples of gold and generated rationales. The most grammatical model is \textbf{boldfaced} and the model that produces the most plausible rationales (according to the evaluation in Table \ref{tab:results_plausibility_with_image}; \sect{sec:plausibility_results}) is \underline{underlined}.}
\label{tab:grammaticality}
\end{table*}

%% file: tables/appendix_fidelity_breakdown.tex
\begin{table*}[t]
\centering
\resizebox{0.8\textwidth}{!}{
\begin{tabular}{l@{\hskip 2mm}l@{\hskip 4mm}l@{\hskip 4mm}lcccc}
\toprule
 &  &&  \textbf{\color{blue} \vcr} & Fidelity & Entity Fidelity & Entity Detail Fidelity& Action Fidelity \\
\midrule
 &  && Baseline & 61.07 & 75.32 & 65.88 & 61.36 \\
 \midrule
\multirow{6}{*}{\rotatebox{90}{\textsc{Rationale}$^{\textsc{VT}}$}} & \multirow{6}{*}{\rotatebox{90}{\textsc{Transformers}}} & \multirow{3}{*}{\rotatebox{90}{\small\uniform}} & Object labels & 60.25 & 77.45 & 69.29 & 66.67 \\
&  & & Situation frames & 62.43 & 77.70 & 66.49 & 61.54 \\
&  & & \visualcomet text inferences & 70.22 & \textbf{79.91} & \textbf{75.74} & 69.63 \\
\cmidrule{3-8}
&  &\multirow{3}{*}{\rotatebox{90}{\small\hybrid}} & Object regions & 54.37 & 73.86 & 58.50 & 59.36 \\
&  & & Situation frames & 54.92 & 73.88 & 62.22 & 60.80 \\
&  & & \visualcomet embeddings & \textbf{72.81} & 79.89 & 75.25 & \textbf{74.41} \\
\midrule
&  & & Human (estimate) & 91.67 & 94.79 & 93.60 & 91.58\\
\bottomrule
\end{tabular}}\vspace{0.2cm}
\resizebox{0.8\textwidth}{!}{
\begin{tabular}{l@{\hskip 2mm}l@{\hskip 4mm}l@{\hskip 4mm}lcccc}
\toprule
 &  && \textbf{\color{blue} \esnlive (contradiction)} & Fidelity & Entity Fidelity & Entity Detail Fidelity& Action Fidelity \\
\midrule
 &  &  & Baseline & 44.74 & 73.21 & 65.05 & 52.19 \\
 \midrule
\multirow{6}{*}{\rotatebox{90}{\textsc{Rationale}$^{\textsc{VT}}$}} & \multirow{6}{*}{\rotatebox{90}{\textsc{Transformers}}} & \multirow{3}{*}{\rotatebox{90}{\small\uniform}} & Object labels & 58.56 & 78.23 & 68.27 & 70.03 \\
 &  &  & Situation frames & \textbf{66.07} & \textbf{82.52} & 71.72 & 71.11 \\
 &  &  & \visualcomet text inferences & 55.26 & 79.24 & 72.00 & \textbf{73.65} \\
\cmidrule{3-8}
 &  & \multirow{3}{*}{\rotatebox{90}{\small\hybrid}} & Object regions & 61.86 & 82.08 & 73.33 & 65.56 \\
 &  &  & Situation frames & 56.16 & 79.87 & 68.78 & 64.29 \\
 &  &  & \visualcomet embeddings & 54.05 & 77.37 & \textbf{79.00} & 62.91 \\
 \midrule
 &  &  & Human (estimate) & 68.17 & 83.07 & 80.85 & 72.71\\
\bottomrule
\end{tabular}}\vspace{0.2cm}
\resizebox{0.8\textwidth}{!}{
\begin{tabular}{l@{\hskip 2mm}l@{\hskip 4mm}l@{\hskip 4mm}lcccc}
\toprule
 &  & & \textbf{\color{blue} \esnlive (entailment)} & Fidelity & Entity Fidelity & Entity Detail Fidelity& Action Fidelity \\
\midrule
 &  &  & Baseline & 74.34 & 82.99 & 93.08 & 94.59 \\
 \midrule
\multirow{6}{*}{\rotatebox{90}{\textsc{Rationale}$^{\textsc{VT}}$}} & \multirow{6}{*}{\rotatebox{90}{\textsc{Transformers}}} & \multirow{3}{*}{\rotatebox{90}{\small\uniform}} & Object labels & 67.39 & 84.31 & 93.46 & 95.59 \\
 &  &  & Situation frames & 72.90 & 84.69 & 92.77 & 95.05 \\
 &  &  & \visualcomet text inferences & 73.14 & 82.66 & 94.77 & 99.55 \\
\cmidrule{3-8}
 &  & \multirow{3}{*}{\rotatebox{90}{\small\hybrid}} & Object regions & 74.34 & \textbf{86.28} & \textbf{95.00} & 96.75 \\
 &  &  & Situation frames & 70.50 & 84.77 & 92.78 & 95.83 \\
 &  &  & \visualcomet embeddings & \textbf{81.53} & 85.60 & 94.65 & \textbf{99.10} \\
\midrule
 &  &  & Human (estimate) & 88.49 & 94.81 & 90.11 & 93.50\\
\bottomrule
\end{tabular}}\vspace{0.2cm}
\resizebox{0.8\textwidth}{!}{
\begin{tabular}{l@{\hskip 2mm}l@{\hskip 4mm}l@{\hskip 4mm}lcccc}
\toprule
 &  && \textbf{\color{blue} \vqae} & Fidelity & Entity Fidelity & Entity Detail Fidelity& Action Fidelity \\
\midrule
 &  &  & Baseline & 52.40 & 74.44 & 74.24 & 67.20 \\
\multirow{6}{*}{\rotatebox{90}{\textsc{Rationale}$^{\textsc{VT}}$}} & \multirow{6}{*}{\rotatebox{90}{\textsc{Transformers}}} & \multirow{3}{*}{\rotatebox{90}{\small\uniform}} & Object labels & 63.47 & 83.84 & \textbf{84.34} & 78.14 \\
 &  &  & Situation frames & 61.07 & 81.82 & 78.52 & 73.85 \\
 &  &  & \visualcomet text inferences & 64.27 & 77.71 & 71.49 & 66.18 \\
 \cmidrule{3-8}
 &  & \multirow{3}{*}{\rotatebox{90}{\small\hybrid}} & Object regions & 69.87 & 86.98 & 79.08 & \textbf{84.75} \\
 &  &  & Situation frames & \textbf{71.47} & \textbf{89.04} & 78.75 & 80.87 \\
 &  &  & \visualcomet embeddings & 60.27 & 77.40 & 76.72 & 64.58 \\
\midrule
 &  &  & Human (estimate) & 89.20 & 94.92 & 94.21 & 92.67\\
\bottomrule
\end{tabular}
}
\caption{\modelname visual fidelity with respect to extracted nouns (entity fidelity), noun phrases (entity detail fidelity), and verbs phrases (action fidelity).}
\label{tab:appendix_fidelity_breakdown}
\end{table*}

%% file: tables/appendix_similarites_between_q_a_and_r.tex
\begin{table*}[t]
\centering
\resizebox{0.7\textwidth}{!}{
\begin{tabular}{llcccc}
\toprule
 &  & \textbf{\vcr} & \makecell[c]{\textbf{\esnlive}\\\textbf{(contradict.)}} & \makecell[c]{\textbf{\esnlive}\\\textbf{(entail.)}} & \textbf{\vqae} \\
 \midrule
 \multirowcell{7}[0pt][l]{Question or\\Hypothesis} & BLEU-1 & 20.25 & 32.57 & 37.71 & 13.49 \\
 & BLEU-2 & 9.78 & 23.29 & 32.93 & 5.69 \\
 & BLEU-3 & 6.48 & 15.92 & 29.59 & 2.46 \\
 & BLEU-4 & 4.58 & 10.94 & 26.83 & 0.97 \\
 & METEOR & 14.05 & 30.25 & 38.47 & 13.13 \\
 & ROUGE-L & 19.64 & 37.45 & 42.93 & 15.44 \\
 & Content Word Overlap & 23.22 & 53.81 & 48.11 & 18.96 \\
\midrule
\multirow{7}{*}{Answer} & BLEU-1 & 27.67 &  &  & 4.96 \\
 & BLEU-2 & 19.07 &  &  & 1.50 \\
 & BLEU-3 & 12.97 &  &  & 0.49 \\
 & BLEU-4 & 9.83 &  &  & 0.00 \\
 & METEOR & 20.22 &  &  & 13.38 \\
 & ROUGE-L & 31.62 &  &  & 10.07 \\
 & Content Word Overlap & 30.09 &  &  & 11.66 \\
 \bottomrule
\end{tabular}
}
\caption{Similarity between question and \textbf{generated} rationale (upper part) and similarity between answer and \textbf{generated}  rationale (lower part). For each dataset, we use rationales from the best \modelname (according to results in Table \ref{tab:results_plausibility_with_image}; \sect{sec:plausibility_results}), except for \esnlive for which we use \uniform fusion of situation frames instead of object labels, because they have comparable plausibility, but situation frames result in better fidelity. We use this model for both \esnlive parts. We use the same samples of data as in the main evaluation.}
\label{tab:appendix_similarities_generated_rationale}
\end{table*}

\begin{table*}[t]
\centering
\resizebox{0.7\textwidth}{!}{
\begin{tabular}{llcccc}
\toprule
 &  & \textbf{\vcr} & \makecell[c]{\textbf{\esnlive}\\\textbf{(contradict.)}} & \makecell[c]{\textbf{\esnlive}\\\textbf{(entail.)}} & \textbf{\vqae} \\
 \midrule
 \multirowcell{7}[0pt][l]{Question or\\Hypothesis} & BLEU-1  & 11.66 & 31.01 & 33.14 & 10.10 \\
 & BLEU-2 & 5.20 & 19.76 & 24.09 & 3.45 \\
 & BLEU-3 & 3.37 & 12.91 & 18.39 & 1.27 \\
 & BLEU-4 & 2.36 & 7.99 & 14.15 & 0.56 \\
 & METEOR & 11.49 & 24.69 & 27.19 & 11.44 \\
 & ROUGE-L & 13.88 & 37.33 & 41.02 & 12.07 \\
 & Content Word Overlap & 13.68 & 47.70 & 43.95 & 14.38 \\
\midrule
\multirow{7}{*}{Answer} & BLEU-1 &15.29 &  &  & 4.00 \\
 & BLEU-2 & 8.13 &  &  & 0.69 \\
 & BLEU-3 & 4.16 &  &  & 0.00 \\
 & BLEU-4 & 2.29 &  &  & 0.00 \\
 & METEOR & 16.35 &  &  & 11.16 \\
 & ROUGE-L & 19.87 &  &  & 8.47 \\
 & Content Word Overlap & 18.01 &  &  & 9.26\\
 \bottomrule
\end{tabular}
}
\caption{Similarity between question and \textbf{gold} rationale (upper part) and similarity between answer and \textbf{gold}  rationale (lower part). We use the same samples of data as in the main evaluation.}
\label{tab:appendix_similarities_gold_rationale}
\end{table*}

%% file: figs/appendix_analysis.tex
\begin{figure*}[!ht]
\centering
\begin{subfigure}[t]{\textwidth}
 \centering
\includegraphics[width=0.55\textwidth]{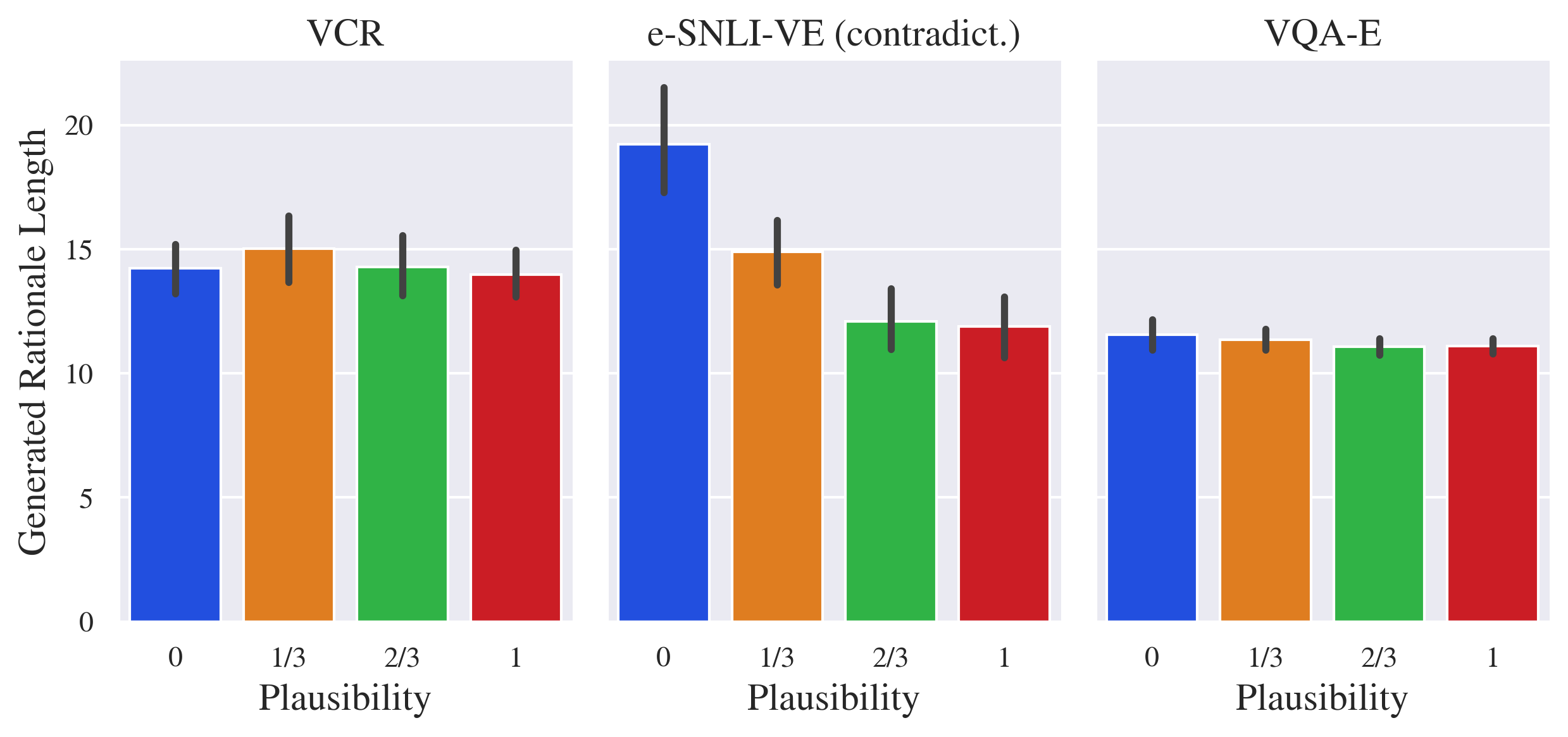}
\caption{The mean and variance of the \textbf{length of generated rationale} with respect to visual plausibility of \textbf{generated rationales}.  The length of generated rationales is similar for plausible and implausible rationales, with exception of \esnlive for which implausible rationales tend to be longer.}
\label{fig:plausibility_vs_gen_rationale_len}
\end{subfigure}\text{ }
\begin{subfigure}[t]{\textwidth}
\centering
\includegraphics[width=0.55\textwidth]{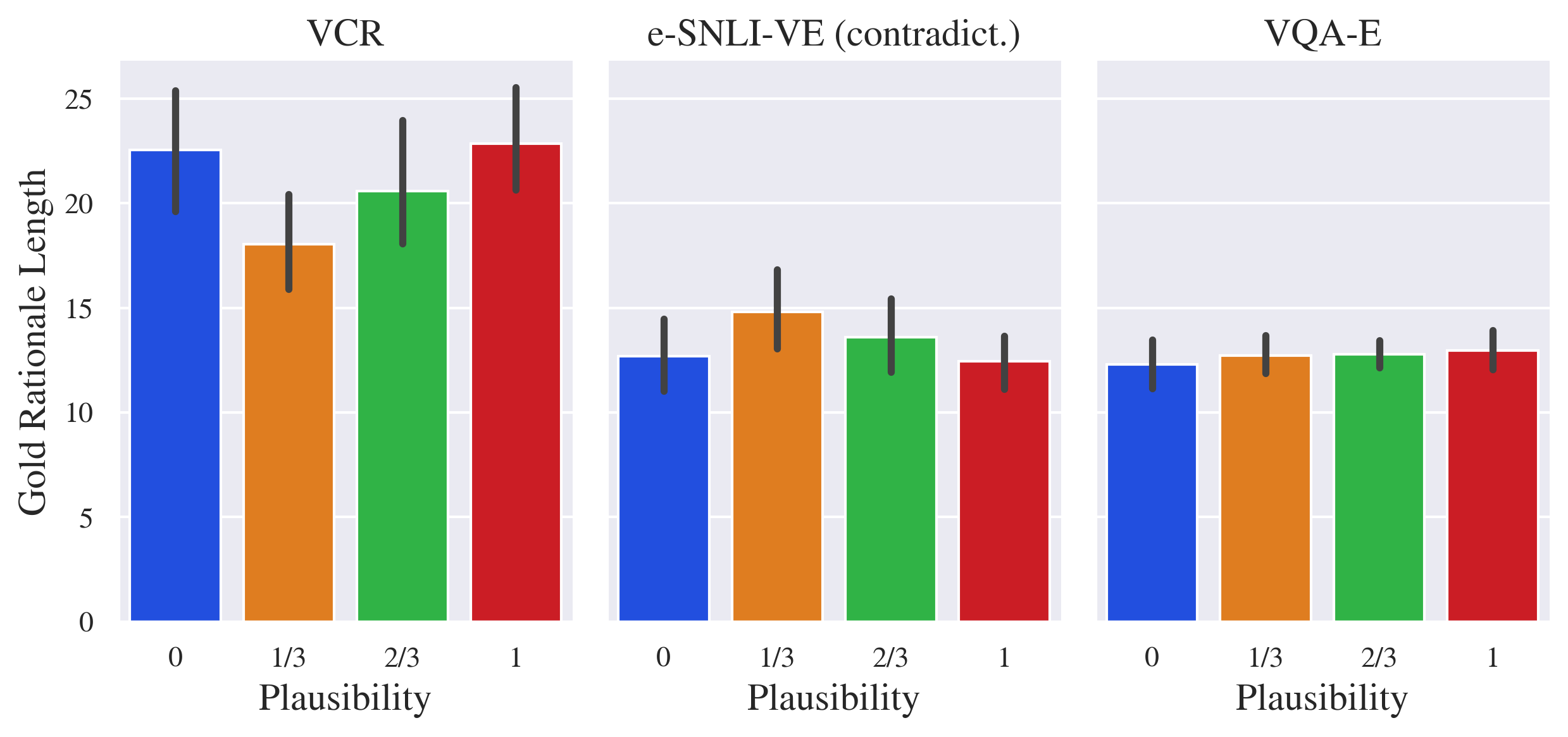}
\caption{The mean and variance of the \textbf{length of gold rationale} with respect to visual plausibility of \textbf{generated rationales}. Rationale generation is not affected by gold rationale length.}
\label{fig:plausibility_vs_gold_rationale_len}
\end{subfigure}\text{ }
\begin{subfigure}[t]{\textwidth}
\centering
\includegraphics[width=0.55\textwidth]{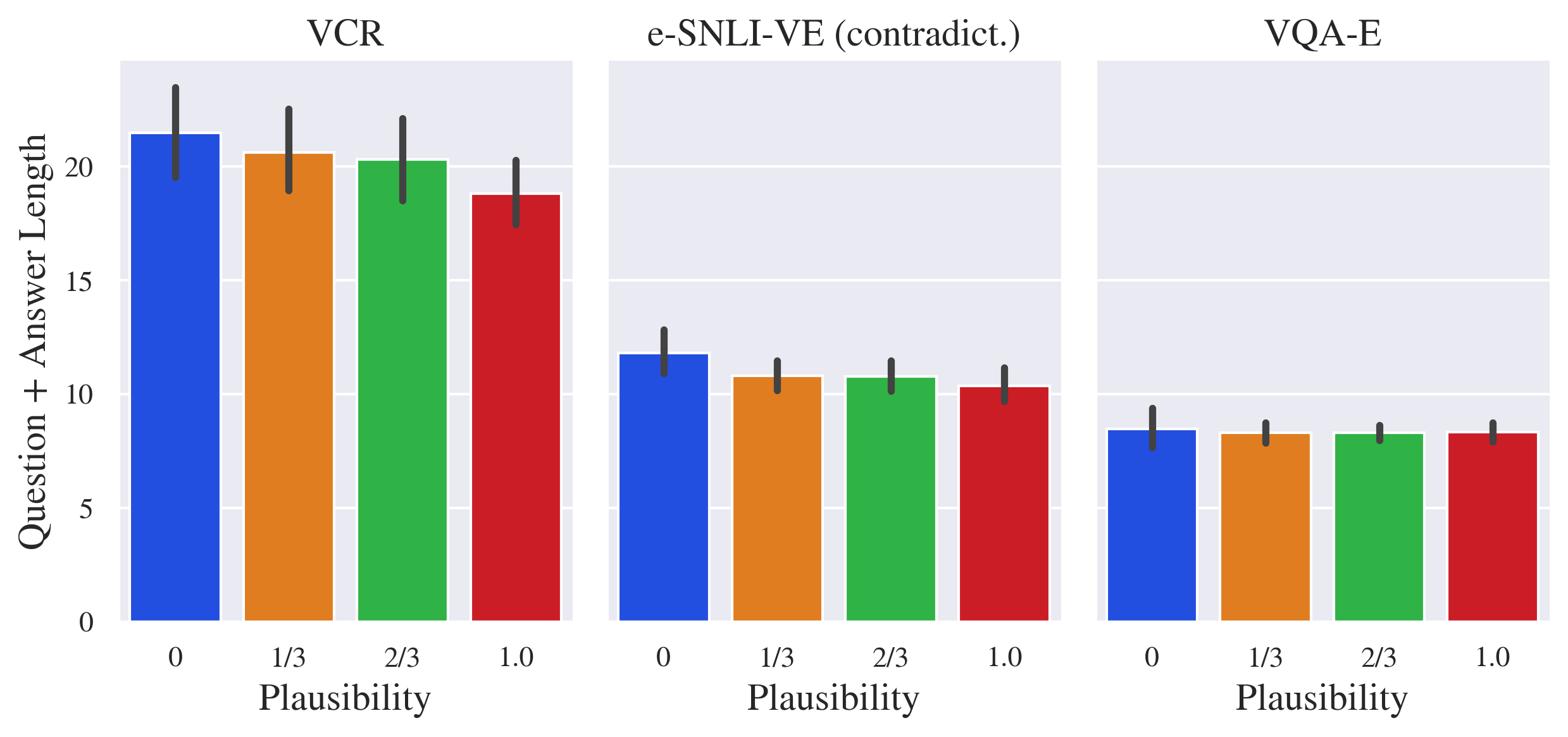}
\caption{The mean and variance of the \textbf{merged question and answer} or just \textbf{hypothesis} with respect to visual plausibility of \textbf{generated rationales}. Plausible rationale tend to rationalize slightly shorter textual context in \vcr and \esnlive.}
\label{fig:plausibility_vs_qa_length_models}
\end{subfigure}\text{ }
\begin{subfigure}[t]{\textwidth}
\centering
\includegraphics[width=0.55\textwidth]{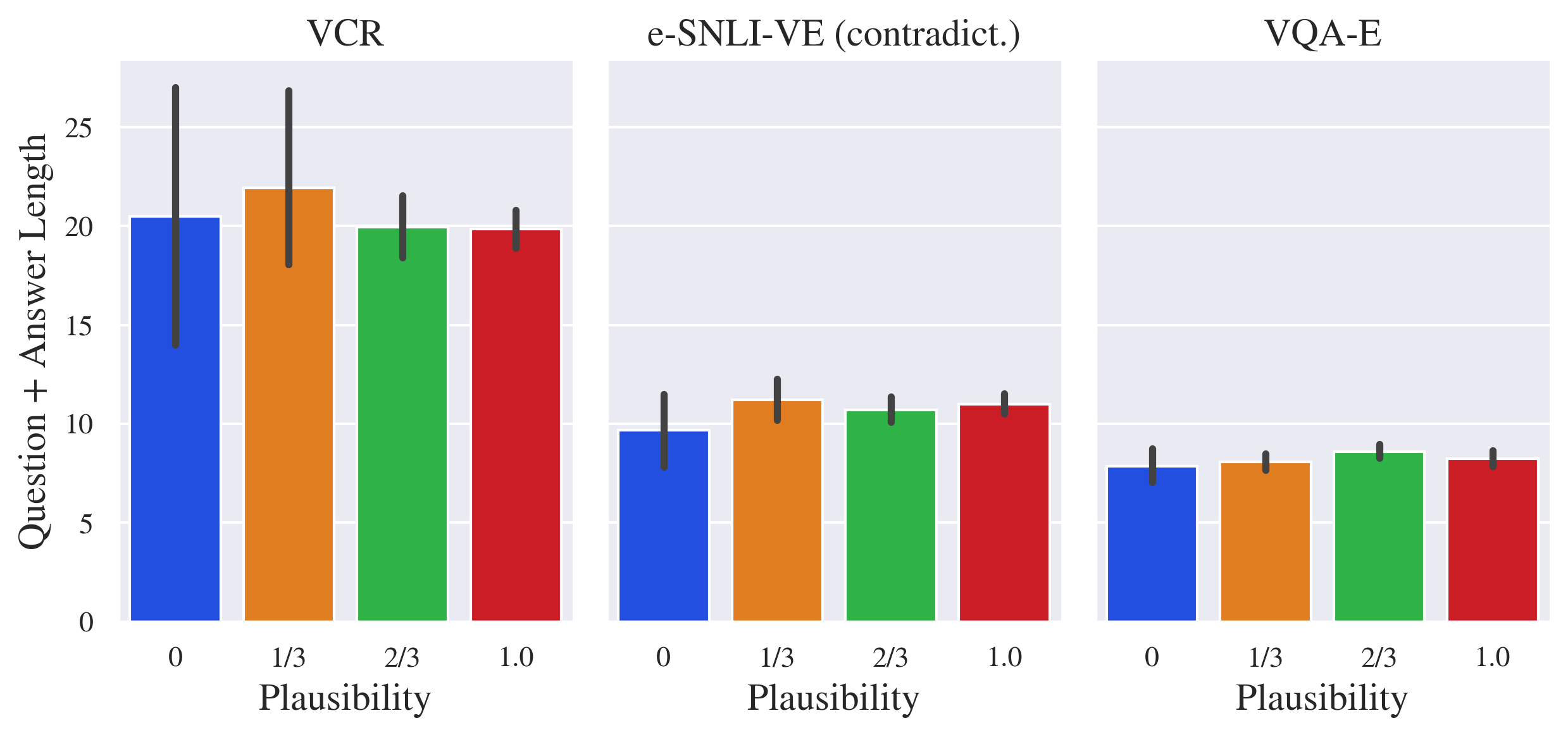}
\caption{The mean and variance of the \textbf{merged question and answer} or just \textbf{hypothesis} with respect to visual plausibility of \textbf{gold rationales}. The small number of implausible \vcr examples also tend to rationalize slightly longer textual contexts, in contrast to \esnlive.}
\label{fig:plausibility_vs_qa_length_human}
\end{subfigure}
\caption{Analysis of plausibility of rationales with respect to input length. Plausibility value is 0 for unanimously implausible, 1 for unanimously plausible, 1/3 for majority vote for implausible, and 2/3 for majority vote for plausible. For each dataset in \ref{fig:plausibility_vs_gen_rationale_len}--\ref{fig:plausibility_vs_qa_length_models}, we use rationales from the best \modelname (according to results in Table \ref{tab:results_plausibility_with_image}; \sect{sec:plausibility_results}), except for \esnlive for which we use \uniform fusion of situation frames instead of object labels, because they have comparable plausibility, but situation frames result in better fidelity. We use this model for both \esnlive parts. We use the same samples of data as in the main evaluation.}
\label{fig:appendix_analysis_input_lenght}
\end{figure*}

\begin{figure*}[t]
\begin{subfigure}[t]{\textwidth}
\centering
\includegraphics[width=0.7\textwidth]{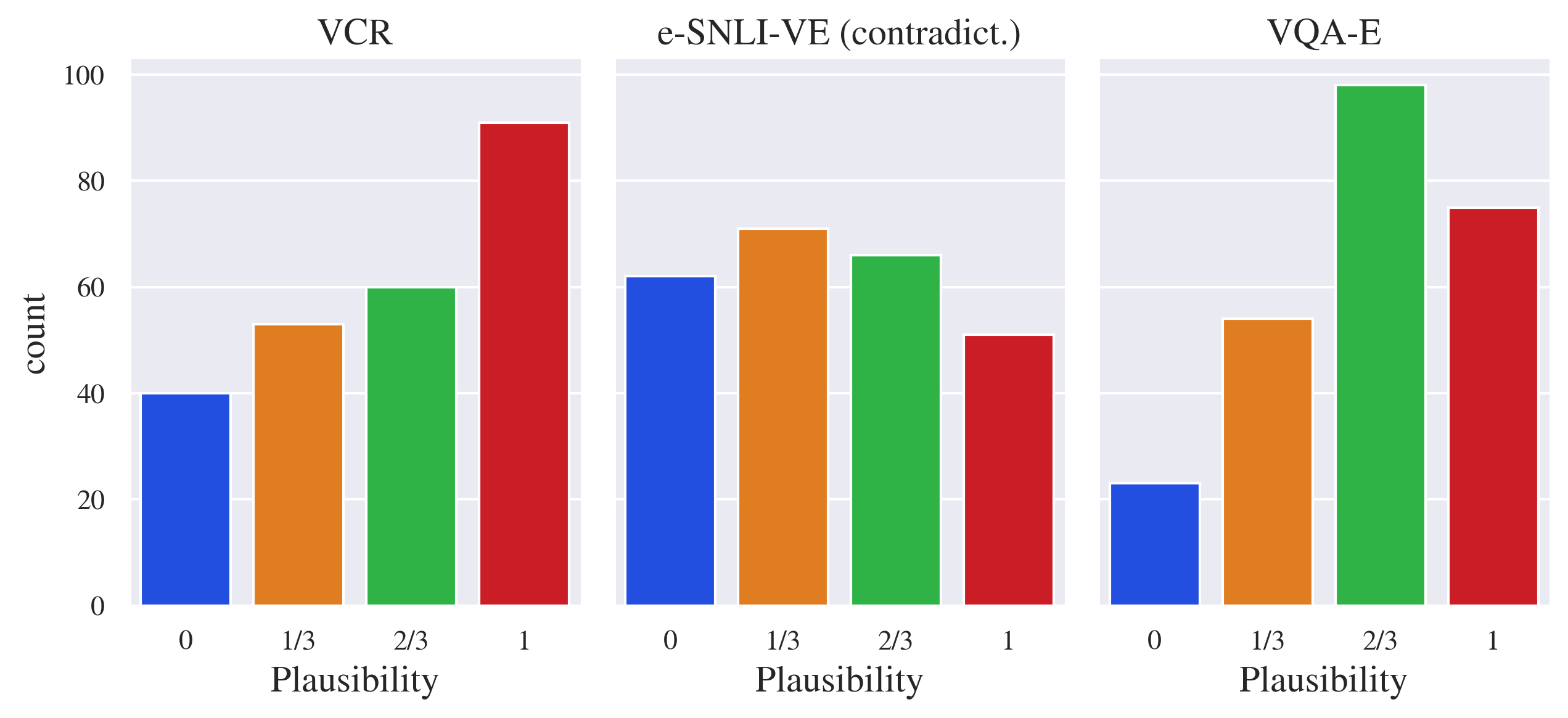}
\caption{Plausibility variation for \textbf{generated} rationales. For each dataset, we use rationales from the best \modelname (according to results in Tables \ref{tab:results_plausibility_with_image}; \sect{sec:plausibility_results}), except for \esnlive for which we use \uniform fusion of situation frames instead of object labels, because they have comparable plausibility, but situation frames result in better fidelity.}
\label{fig:generations_plausibility_variation}
\end{subfigure}\text{ }
\begin{subfigure}[t]{\textwidth}
\centering
\includegraphics[width=0.7\textwidth]{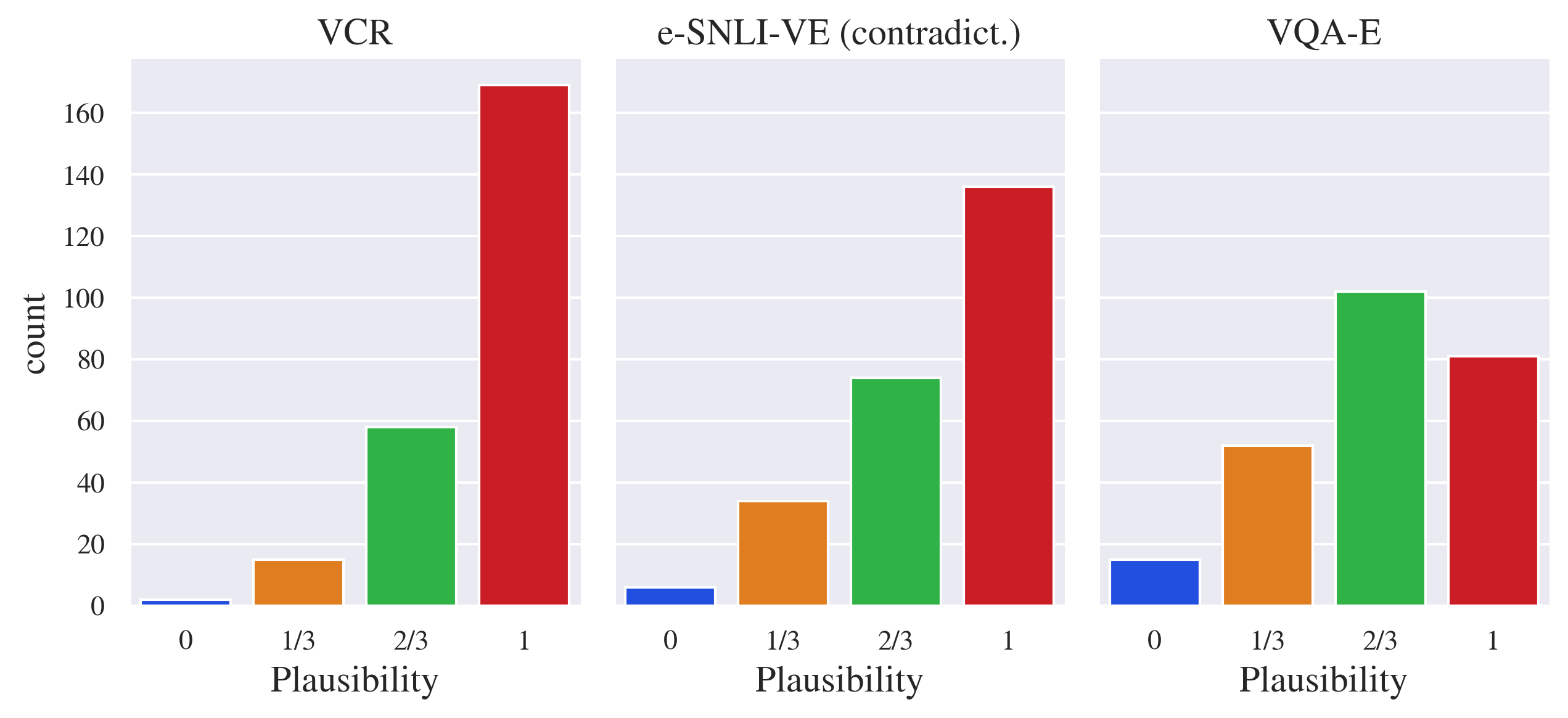}
\caption{There is less variation for \textbf{gold} rationales.}
\end{subfigure}
\caption{Analysis of variation of plausibility judgments. Plausibility value is 0 for unanimously implausible, 1 for unanimously plausible, 1/3 for majority vote for implausible, and 2/3 for majority vote for plausible. We use the same samples of data as in the main evaluation.}
\label{fig:appendix_plausibility_variation}
\end{figure*}